\documentclass[11]{interact}
\usepackage{caption, subcaption, graphicx}
\usepackage[utf8]{inputenc}

\usepackage{epsfig,graphicx,amsmath,amssymb,footmisc,xspace,float,bm,enumerate,mathrsfs,verbatim,multirow}

\usepackage{epstopdf} 
 \usepackage[numbers,sort&compress]{natbib}
\bibpunct[, ]{[}{]}{,}{n}{,}{,}
 \makeatletter
\def\NAT@def@citea{\def\@citea{\NAT@separator}}
\makeatother
 \usepackage{booktabs}
\usepackage{multirow,bigdelim}
\usepackage{booktabs}
\usepackage{siunitx}

\theoremstyle{plain}

\theoremstyle{definition}

\theoremstyle{remark}

 \newcommand{\rb}[1]{\footnotesize{#1}}

\newcommand\beq{\begin{equation}}
\newcommand\eeq{\end{equation}}
\newcommand\btheta{\boldsymbol{\theta}}

\def\bxi{\boldsymbol{\xi}}
\def\bM{\boldsymbol{M}}
\def\hlambda{\hat{\lambda}}
\def\tlambda{\tilde{\lambda}}

\def\tbM{\tilde{\bM}}
\def\bN{\boldsymbol{N}}
\def\bU{\boldsymbol{U}}
\def\bD{\boldsymbol{D}}

\def\tbN{\tilde{\bN}}
\def\btheta{\boldsymbol{\theta}}
\def\hbtheta{\hat{\btheta}}
\def\tbtheta{\tilde{\btheta}}

\def\summ{\sum_{i=1}^2\sum_{j=1}^{n_i}}

\def\txi{\tilde{\boldsymbol{\xi}_{ij}}}
\def\minv{\tilde{\mathbf{M}}^-}
\def\btth{\tilde{\boldsymbol{\theta}}}

\def\bt{\boldsymbol{\theta}}

\def\bM{\mathbf{M}}

\def\bN{\mathbf{N}}
\def\bV{\mathbf{V}}

\def\bD{\mathbf{D}}

\def\cF{\mathcal{F}}

\def\bT{\mathbf{T}}

\def\E{\mathbb{E}}
\def\bxi{\boldsymbol{\xi} }

\def\tbxi{\tilde{\bxi} }
\usepackage{setspace}
\begin{document}

 \title{New Methods for Detecting Concentric Objects With High Accuracy }\author{\name{Ali A.~Al-Sharadqah\textsuperscript{a}\thanks{Email: ali.alsharadqah@csun.edu} and Lorenzo Rulli\textsuperscript{b}\thanks{Email: l.rulli@sbcglobal.net}}\affil{\textsuperscript{a}Department of Mathematics, California State University--Northridge, CA 91330, US; }}

\maketitle

\begin{abstract}
Fitting concentric geometric objects to digitized data is an important problem in many areas such as iris detection, autonomous navigation, and industrial robotics operations. There are two common approaches to fitting geometric shapes to data: the geometric (iterative) approach and algebraic (non-iterative) approach. The geometric approach is a nonlinear iterative method that minimizes the sum of the squares of Euclidean distances of the observed points to the ellipses and regarded as the most accurate method, but it needs a good initial guess to improve the convergence rate. The algebraic approach is based on minimizing the algebraic distances with some constraints imposed on parametric space. Each algebraic method depends on the imposed constraint, and it can be solved with the aid of the generalized eigenvalue problem. Only a few methods in literature were developed to solve the problem of concentric ellipses. Here we study the statistical properties of existing methods by firstly establishing a general mathematical and statistical framework for this problem. Using rigorous perturbation analysis, we derive the variances and biasedness of each method under the small-sigma model. We also develop new estimators, which can be used as reliable initial guesses for other iterative methods. Then we compare the performance of each method according to their theoretical accuracy. Not only do our methods described here outperform other existing non-iterative methods, they are also quite robust against large noise. These methods and their practical performances are assessed by a series of numerical experiments on both synthetic and real data.
 
\end{abstract}

\begin{keywords}
 multiple concentric ellipses, non-iterative methods, perturbation analysis, mean-square error, bias removal.
\end{keywords}


One of the most fundamental tasks in computer vision and image processing is determining the 2 or 3-dimensional shapes of objects, from images. This problem is known by \emph{geometric estimation} problem. It has a vital role in computer vision, biomedical imaging, archaeology, agriculture, nuclear physics, and many other fields. In biometric security, the objective is to correctly identify an individual based on his or her  physical features. One of the fastest and most effective approaches for identifying an individual is through scanning the person’s iris, located in eye. After taking a photographic image of a person’s eyes, computers use edge detection techniques to extract digitized observations from pixels. Geometric estimation is then used to determine the person’s iris, in particular the oval shapes within the iris. Mathematically, the main problem can be formalized as follows: \emph{Given a set of digitized observations extracted from the inner and the outer concentric ellipses, we want to determine the concentric ellipses}. This problem lies under the umbrella of geometric estimation. In the problem of geometric estimation, one focuses on determining the 2- or 3-dimensional shapes of objects from images. Therefore, it is one of the most fundamental tasks in computer vision and image processing \cite{Bennamou,HartleyZimm03,KanataniSug2016b}. It also has a vital role in petroleum engineering \cite{Heidari93,KST11}, astronomy \cite{AX10,ZYC11}, bioinformatics \cite{Kummer18}, and object classifications \cite{Connell01}.

While the problem of fitting a single ellipse to data has been studied intensively in the literature \cite{Prasad2013,Wailbel2016}, the research on the problem of fitting multiple concentric objects to digitized data is still immature. In this paper, we assume that $n_i$ digitized observations were distributed around the $i^{ th}$ elliptical arcs, for each $i=1\ldots, K$, where $K$ is the number of concentric ellipses that have a common center and a common tilt angle. Since in iris detection $K=2$, we will use that in our analysis. This eases our formulation, while its extension for $K>2$ will be justified at the end of Section \ref{NumeExpeConc}.

\begin{figure}[H]
\centering
  \includegraphics[width=.3\textwidth,height=.25\textheight,keepaspectratio]{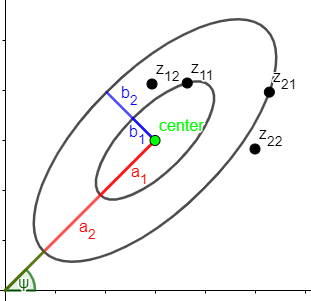}
 \vspace{-3mm} \caption{Geometric parameters of concentric ellipses with common center $(x_c,y_c)$ and tilt angle $\psi$. $(a_1,b_1)$ are the semi-major and semi-minor axes of the inner ellipse. $(a_2,b_2)$ are the semi-major and semi-minor axes of the outer ellipse. The point $\mathbf{z}_{11}=(x_{11},y_{11})$ lies on the inner ellipse, while the point $\mathbf{z}_{21}=(x_{21},y_{21})$ lies on the outer ellipse.}
  \label{fig:concentricellipseparam}
  \end{figure}
As depicted in Figure \ref{fig:concentricellipseparam}, our problem becomes how to compute the common center $(x_c,y_c)$, the common tilt angle $\psi$, the semi-major axes $a_1$ and $a_2$, and the semi-minor axes $b_1$ and $b_2$ for both inner and outer concentric ellipses, respectively. In this paper, we assume that the semi-major and semi-minor axes of the outer ellipse are scaled multiples of the inner ellipse axes, i.e., $a_2=a_1(1+\varepsilon)$, $b_2=b_1(1+\varepsilon)$. Our task becomes to estimate the \emph{geometric parameter vector} $\boldsymbol{\phi}=(x_c,y_c,a_1,b_1,a_2,b_2,\psi)^\top$. However, dealing with the geometric parameters leads to a highly nonlinear problems due to the nonlinear implicit relationship between the point $\boldsymbol{z}$ and $\boldsymbol{\phi}$. Alternatively, researchers in this situation avoid using $\boldsymbol{\phi}$. In our problem, the following linearization produces the so-called the \emph{algebraic parameter vector} $\bt = (A,B,C,D,E,F_1,F_2)^\top$. Dealing with $\bt$ reduces the complexity of the problem significantly. Then after $\bt$ is estimated, one can retrieve the geometric parameters $\bt$ through the relationship of the two parametric spaces, the conversion formulas between them are justified in Appendix \ref{relation}. Note here that $\btheta\in \mathbb{R}^{7}$ due to  the common center and title angle. To justify this, let $\mathbf{z}_{11}=(x_{11},y_{11})$ and $\mathbf{z}_{21}=(x_{21},y_{21})$ be two points that lie on the inner and the outer ellipses, respectively, then the point $\mathbf{z}_{i1}$ lies on the $i^{th}$ ellipse, then 
\begin{equation}\label{Eq00}
Ax_{i1}^2 + 2Bx _{i1} y_{i1} + Cy_{i1}^2 + 2f_0Dx_{i1} + 2f_0E y_{i1} + f_0^2F_i=0\, , \quad \text{ for  } i=1,2,
\end{equation}
where the positive constant $f_0$\footnote{Our experiments in this paper used images with a $320\times 290$ resolution. Therefore, the extracted data was normalized using $f_0=100$ pixels so that the transformed data of order 1; see \cite{KanataniSug2016b} for more details. } is a scale factor that is used to normalize data, and as such, the transformed data stays within a reasonable numerical range. It is also worth mentioning here that sharing the same center and the tilt angle implies that the algebraic parameters $\boldsymbol{\theta}_1=(A,B,C,D,E)^{\top}$ are also common for both ellipses. i.e., the two ellipses differ in only the constant term $F_i$, $i=1,2$. The mathematical details of Eq. \eqref{Eq00} are deferred to Appendix \ref{ApxGeoCon}.

In passing, the three commonly approaches used to the problem of fitting a \emph{single} ellipse to data are the (iterative) \emph{Orthogonal Distance Regression method} (ODR) \cite{AliHo0,ChangWeiduo19}, the (iterative) \textit{Gradient Weighted Algebraic} (GRAF)\cite{Chojnacki2} and  the (iterative) \textit{Renormalization Method} \cite{Kanatani1}, as well as many other non-iterative algebraic methods \cite{AliHo0,Chernov1,Chernov2}. The ODR is highly nonlinear iterative method that can be obtained by minimizing the sum of the squares of Euclidean distances of the observed points to the ellipse. Although it is regarded as the most accurate fitting method, it is computationally expensive and needs a good initial guess that help to improve its convergence rate. A Less expensive iterative method but still falling behind the ODR is the GRAF. 

The drawback of the iterative methods is that they need iterative algorithms, and as such, they are computationally expensive. Being iterative, they are prone to divergence when the data is contaminated with a large levels of noise, clustered along short arcs, or seeded by a poor initial guess. Therefore, providing a good initial guess will reduce the divergence issue. The initial guess will be obtained by non-iterative algebraic methods that minimize the sum of the squares of the so-called algebraic distances. Each algebraic method impose a constraint on the parametric space in order to get a nontrivial solution and eliminate any indeterminacy issue. 

This paper is devoted to study the non-iterative algebraic methods for the problem of multiple concentric ellipses to data. In this paper we discuss the recent existing methods and we express all methods in general mathematical framework which not only allows us to understanding the common properties between all method but also developing new ones. We also apply a comprehensive comparison between all methods, and as such understanding how each method differs from others in practice. 
Our analysis is assessed by a series of numerical experiments that confirm the superior performance of our methods. Our results are also validated using real data extracted from images of human eyes. This paper is structured as follows. Section \ref{Intro} introduces the existing algebraic methods and discusses their numerical implementations. It also explains how the so-called Taubin method in ellipse fitting can be extended to the problem of multiple concentric ellipses. Section \ref{ErrorAnalysis} discusses the statistical properties of the three methods. Section \ref{compare} discusses the accuracy of three methods and compares between between them. Section \ref{HYPER} introduces two new methods and discusses their practical implementations. Section \ref{NumeExpeConc} validates our results through a series of numerical experiments on both synthetic and real data. Finally, Section \ref{Conclusion} concludes our results, while Section \ref{Apx} is devoted to being an appendix for our technical details.

 \section{Introduction and Background}\label{Intro}
There are only two existing noninterative methods that have been devloped recently to the problem of concentric ellipses: the standard LS by Ma and Ho \cite{MaHo} and the O'Leary method by O'Leary \emph{et al.} \cite{O'Leary2004}. In this section we will go over those methods and we will also extend the so-called Taubin's fit \cite{Taubin}, a widely known method in geometric estimation. It is worth mentioning here that we can not directly implement the Taubin method due to the some singularity issues in this extension. To eliminate the singularity problem, some necessary modifications will be implemented. To compensate with, we first assume that $\left\{ (x_{ij},y_{ij})\right\}_{j=1}^{n_i}$ are the $n_i$ observed points distributed along the $i^{ th}$ ellipse, then an algebraic method shall minimize
\begin{equation} \label{J0}
{\rb J(\boldsymbol{\theta})=
\sum_{i=1}^2 \sum_{j = 1}^{n_i}(Ax_{ij}^2 + 2Bx_{ij} y_{ij} + Cy_{ij}^2 + 2f_0Dx_{ij} + 2f_0Ey_{ij} + f_0^2F_i)^2\rightarrow \min}
\end{equation}
and define the so-called the \emph{carrier vector} by
$
\boldsymbol{\xi}^\top=(x^2,2xy,y^2,2f_0x,2f_0y,f_0^2\hat{\delta}_{i1},f_0^2 \hat{\delta}_{i2})
$
where $\hat{\delta}_{ik}$ is the Kronecker Delta function (i.e., $\hat{\delta}_{ik}=1$ if $i=k$, and 0 elsewhere). More explicitly, the carrier for the inner ellipse ($i=1$) and the outer ellipse ($i=2$) can be written as $\boldsymbol{\xi_{11}} = (x_{11}^2,2x_{11}y_{11},y_{11}^2,2x_{11},2y_{11},1,0)^\top,$ and $
\boldsymbol{\xi_{21}} = (x_{21}^2,2x_{21}y_{21},y_{21}^2,2x_{21},2y_{21},0,1)^\top$, 
which in vector notations can be expressed as $\boldsymbol{\xi}_{i1}^\top\bt=0$, for each $i=1,2$. Therefore, Eq. \eqref{J0} can be expressed in matrix forms as
\begin{equation}\label{J}
J(\boldsymbol{\theta}) =\sum_{i=1}^2 \sum_{j = 1}^{n_i}\btheta^\top\left ( \boldsymbol{\xi}_{ij}\boldsymbol{\xi}_{ij}^\top\right)\btheta: =\bt^\top\bM\bt,
\end{equation}
where $ \boldsymbol{\xi_{ij}}=(x_{ij}^2,2x_{ij}y_{ij},y_{ij}^2,2f_0x_{ij},2f_0y_{ij},f_0^2\hat{\delta}_{i1},f_0^2\hat{\delta}_{i2})^\top$, for each $i=1,2,$ and
\begin{equation}\label{M}
\bM  = \summ \bM_{ij} = \summ \bxi_{ij}\bxi_{ij}^\top,\quad \bM_{ij}=
\left[
\begin{array}{cc}
\bM_{ij}^{11} & \bM_{ij}^{12}\\
(\bM_{ij}^{12})^\top & \bM_{ij}^{22} \\
\end{array}
\right]
\end{equation}
with
 \begin{equation}
 \resizebox{.98\hsize}{!}{$
\bM_{ij}^{11} =
\left[
\begin{array}{ccccc}
x_{ij}^4 & 2x_{ij}^3y_{ij} &  x_{ij}^2y_{ij}^2 &  2f_0x_{ij}^3 &  2f_0x_{ij}^2 y_{ij}\\
2 x_{ij}^3y_{ij} & 4 x_{ij}^2y_{ij}^2 & 2 x_{ij}y_{ij}^3 & 2f_0 x_{ij}^2y_{ij} & 2 f_0x_{ij}y_{ij}^2 \\
 x_{ij}^2y_{ij}^2 & 2 x_{ij}y_{ij}^3 &  y_{ij}^4 & 2 f_0x_{ij} y_{ij}^2 & 2 f_0y_{ij}^3 \\
2 f_0x_{ij}^3 & 2f_0 x_{ij}^2y_{ij} &2 f_0 x_{ij}y_{ij}^2 &  4f_0^2x_{ij}^2 & 4 f_0^2x_{ij}y_{ij} \\
2f_0 x_{ij}^2y_{ij} & 2 f_0x_{ij}y_{ij}^2 &  2f_0y_{ij}^3 &4 f_0^2x_{ij}y_{ij} &4f_0^2 y_{ij}^2 \\
\end{array}
\right] , \quad 
\bM_{ij}^{12} =
\left[
\begin{array}{ccccc}
 f_0^2x_{1j}^2 & f_0^2x_{1j}y_{1j}&f_0^2y_{1j}^2 &2 f_0^3x_{1j}&2f_0^3 y_{2j}\\
 f_0^2x_{2j}^2 &2f_0^2x_{2j}y_{2j}  &f_0^2y_{2j}^2 &2f_0^3x_{2j} &
 2f_0^3y_{1j}
\end{array}
\right] ^\top,
$}
\end{equation}
while
$
\bM_{1j}^{22} = {\rm Diag}\left(f_0^4,0\right)$ and $\bM_{2j}^{22} ={\rm  Diag}\left(0,f_0^4\right)$. Since $\bM$ is a positive semi-definite matrix and $J$ is a quadratic function of $\bt$, $J$ is minimized at $\bt=\mathbf{0}$. To avoid the unwanted trivial solution of Eq. \eqref{J}, a quadratic constraint shall be imposed. Such a constraint can be expressed as $\btheta^\top\bN\btheta=1$. Thus, minimizing $J$ turns out to be one of the solutions for the generalized eigenvalue problem (GE) \begin{equation}\label{singleGEP}
\bM\btheta=\lambda\bN\btheta.
\end{equation}

\subsection{The O'Leary method}
Only a few methods in literature were developed to solve the problem of concentric ellipses. O'Leary \emph{et al.} \cite{O'Leary2004} developed fitting methods for coupled geometric objects, including concentric ellipses, by utilizing a quadratic constraint which guarantees coupled ellipses. Since the goal is to estimate the parameters of the conics, O'Leary \emph{et al.} impose a constraint on the parametric space that guarantees an ellipse, i.e., $AC-B^2>0$. Since multiplying this inequality by a nonzero constant does not change the geometric shape, O'Leary \emph{et al.} impose the quadratic constraint $\bt^\top\bN_{ \rm OL}\bt =1$, where the constraint matrix $\bN_{ OL}$ of O'Leary's method is
\begin{equation}\label{OlearyConstraintMatrix}
\resizebox{.6\hsize}{!}{$\bN_{ OL} =\left[\begin{array}{cc}
\mathbf{K}& \mathbf{0}_{3\times 4} \\
\mathbf{0}_{4\times 3} &\mathbf{0}_{4\times 4}
\end{array}\right],\qquad \qquad
\mathbf{K}=
\left[
\begin{array}{ccc}
 0 & 0 & \tfrac{1}{2} \\
 0 & -1 & 0 \\
 \tfrac{1}{2} & 0 & 0
 \end{array}
\right] .$}
\end{equation}
That is, the O'Leary method minimizes $J(\btheta)$ subject to $\ \bt^\top\bN_{ OL}\bt=1$. Since $\bM$ is positive semi-definite, then  this problem is equivalent to minimizing $\bt^T\bM\bt $ subject to $\bt^\top\bN_{\rm OL}\bt =1$. Now, by introducing the Lagrange multiplier $\lambda$, we minimize the function $\bt ^\top\bM\bt -\lambda(\bt ^\top\bN_{\rm OL}\bt-1)$, which again turns out to be a generalized eigenvalue problem $\bM\bt =\lambda\bN_{\rm OL}\bt $. The solution $\hbtheta_{\rm OL}$ for the matrix pair (pencil of matrices) $(\bM,\bN_{\rm OL})$ is the generalized eigenvector corresponding to the smallest (positive) generalized eigenvalue $\lambda$
.
That is, the matrix $\bN_{\rm OL}$ has only one positive generalized eigenvalue and two negative generalized eigenvalues and four zeros, i.e., $ \tfrac{1}{2}$,$-1$,$\tfrac{-1}{2},0$. Also, the matrix $\bM$ is positive definite in general\footnote{The exceptional case happens when all observations lie exactly on the geometric shapes, then $\bM$ is singular. However, this situation in practice appears with zero probability zero.}. Accordingly, the generalized eigenvalues of the pencil $ (\bM, \bN_{\rm OL})$ are all real, exactly two of them are negative, one is positive, and three are zeros. To determine $\hat{\bt}_{\rm OL}={\rm arg}_{\bt}\, \cF_{\rm OL}(\bt)$, we observe that $ \hat{\bt}_{\rm OL} ^\top\bM\hat{\bt}_{\rm OL} = \lambda\hat{\bt}_{\rm OL} ^\top\hat{\bt}_{\rm OL} =\lambda$. This means that $\lambda>0 $ and as such, the O'Leary method $\hat{\bt}_{\rm OL}$ is the generalized eigenvector corresponding to the smallest positive generalized eigenvalue $\lambda$.

\subsection{The Least Square Method}
Ma and Ho \cite{MaHo} use the Least Squares (LS) method to fit concentric circles and ellipses but under more general statistical assumptions \cite{MaHo}. The LS Method is the simplest and fastest method that minimizes $J(\btheta)$ subject to the constraint $ \|\bt\|^2 = 1.$ With imposing this quadratic constraint, one can show the solution of the LS can be obtained by solving the eigenvalue problem $\bM\bt=\lambda\bt$. Since $\bM$ is positive definite (in generic cases), $\lambda$ must be positive as well; therefore, all eigenvalues are positive. That is, if we pre-multiply $\bM\bt=\lambda\bt$ by $\bt^\top$, we obtain $\bt^\top\bM\bt=\lambda\|\bt\|^2$. This means $\lambda=\frac{\bt^\top\bM\bt}{\|\bt\|^2}>0$. Therefore, the LS estimator, $\hbtheta_{\rm LS}$, is the eigenvector corresponding to the smallest positive eigenvalue. It is worth mentioning here that numerical experiments show that the accuracy of the LS becomes questionable when data are clustered on short arcs.

\subsection{The Taubin Method}
The Taubin method \cite{Taubin} is one of the most popular and accurate fitting methods in the problem of fitting geometric shapes and it is data-dependent. Unlike the LS and the O'Leary methods, the Taubin method incorporates all information available in measurement errors of the data. The Taubin method can be extended here to our problem as follows. First we assume in this paper that each data point $\boldsymbol{z}_{ij}$ is a noisy version of the true point $\tilde{\boldsymbol{z}}_{ij}$. The true point lies on the true curve, i.e., the true point satisfies the geometric constraint determined by ellipses. One can express this as $\boldsymbol{z}_{ij}=\tilde{\boldsymbol{z}}_{ij}+\boldsymbol{e}_{ij}$, where $\tilde{\boldsymbol{z}}_{ij}=(\tilde{x}_{ij},\tilde{y}_{ij})$ is the true point of $\boldsymbol{z}_{ij}$ and $\boldsymbol{e}_{ij}$ is the disturbance vector with zero mean and variance $\sigma^2 \mathbf{I}_2$, where $\mathbf{I}_2$ is identity matrix of size 2. Accordingly, this problem lies under the umbrella of \emph{Errors--In--Variables} (EIV) models, which is different and much more difficult than classical regression. Therefore, the carriers $\bxi_{ij}$ are quadratic functions of $(\delta_{ij},\epsilon_{ij})$. Thus, by using the Taylor expansion of $\bxi_{ij}$ about $\tilde{\boldsymbol{z}}_{ij}=(\tilde{x}_{ij},\tilde{y}_{ij})$, one can show that the leading term of its variance matrix is $\bV[\bxi]=\sigma^2\bV_0[\bxi]$ where $\bV_0$ is the normalized covariance matrix. Its formal expression is
\begin{equation}\label{V0}
\resizebox{.55\hsize}{!}{$
\bV_0[\bxi]= 4
\begin{bmatrix}
{x}_{ij}^2 &  {x}_{ij}{y}_{ij} & 0 & f_0{x}_{ij} & 0 & 0 & 0 \\
{x}_{ij}{y}_{ij} &  {x}_{ij}^2 + {y}_{ij}^2 & {x}_{ij}{y}_{ij} & f_0{y}_{ij} & f_0{x}_{ij} & 0 & 0 \\
0 & {x}_{ij}{y}_{ij} & {y}_{ij}^2 & 0 & f_0{y}_{ij} & 0 & 0\\
f_0{x}_{ij} & f_0{y}_{ij} & 0 & f_{0}^2 & 0 & 0 & 0 \\
0 & f_0 {x}_{ij} & f_0{y}_{ij} & 0 & f_{0}^2 & 0 & 0\\
0 & 0 & 0 & 0 & 0 & 0 & 0\\
0 & 0 & 0 & 0 & 0 & 0 & 0\\
\end{bmatrix}.
$}
\end{equation}
If we define the normalization matrix $\rb{\bN_{\rm T}=\sum_{i=1}^2\sum_{j=1}^{n_i} \bV[\xi_{ij}]}$, then the Taubin method minimizes the objective function
\begin{equation}\label{TaubinFit}
    J_{\rm  T}(\bt) = \frac{\summ(\bxi,\bt)^2}{\summ(\bt,\bV_0[\bxi]\bt)} = \frac{\bt^\top\bM\bt}{\bt^\top\bN_{\rm T}\bt}.
\end{equation}
Differentiating $J_{\rm T}$ with respect to $\bt$ yields ${\rb{ \tfrac{\partial}{\partial\bt}J_{ \rm T}(\bt)= \tfrac{1}{\bt^\top\bN_{ \rm T}\bt}\bM\bt - \tfrac{\bt^\top\bM\bt }{(\bt^\top\bN_{\rm T}\bt)^2}\bN_{\rm  T}\bt}}$. Setting the resulting estimating equation to zero yields the new GE problem $\rb{\bM\bt = \lambda\bN_{\rm T}\bt}$, where $\rb{\lambda = \tfrac{\bt^\top\bM\bt}{\bt^\top\bN_{\rm T}\bt}}$. It is worth mentioning here that both $\bM$ and $\bN_{ T}$ are positive semi-definite, hence, $\lambda$ must be non-negative as well. Therefore, the solution of the Taubin method $\hbtheta_{ \rm T}$ is the unit generalized eigenvector corresponding to smallest positive generalized eigenvalue $\lambda$. To implement the Taubin method numerically,  we indirectly solve the problem by decomposing this GE problem into
 $
 \bM^{11}\bt_1+\bM^{12}\bt_2=\lambda\check{\bN}_{\rm T}\bt_1$ and  $\bM^{12\top}\bt_1+\bM^{22}\bt_2=\mathbf{0}$, 
where ${\rb \bM^{kl}=\sum_{i=1}^2\sum_{j=1}^{n_i} \bM_{ij}^{kl}}$ and $\check{\bN}_{\rm T}$ is a square matrix of size 5 that is obtained after removing the last two rows and columns of $\bN_{\rm T}$. This yields ${\rb \hbtheta_2 = -(\bM^{22})^{-1}\bM^{12\top}\hbtheta_1}$. Thus, we can solve the problem by firstly solving ${\rb{ \check{\bM}\hbtheta_1 = \lambda\check{\bN}_{\rm T}\hbtheta_1}}$, where $\check{\bM}= \bM^{11}-\bM^{12}(\bM^{22})^{-1}\bM^{12\top}$, and then we retrieve $\hbtheta_1$, and as such, the Taubin solution is $\rb{\hbtheta_{\rm T}^\top =(\hbtheta_{1}^\top,\hbtheta_2^\top)}$.

\section{Error Analysis}\label{ErrorAnalysis}
Since geometric estimation plays a vital role in computer vision and image processing, our adopted statistical assumptions are tailored to those applications. For instance, we use the \emph{small-sigma model} \cite{Kanatani2011} to study the asymptotic behavior of estimators. That is, an estimator $\hbtheta $ is called consistent in the following sense: $\hbtheta$ is approaching the true value of the parameters $\tbtheta$ in probability as the noise level $\sigma$ is approaching 0. This is regarded as the minimal requirement for any legitimate fitting algorithm. To distinguish this requirement from the traditional consistency, we call estimators satisfying this limiting behavior \emph{geometrically consistent estimators}\footnote{All existing methods discussed here are geometrically consistent.}. Accordingly, we will deal here with geometric consistent estimators $\hbtheta$ that solves the generalized eigenvalue problem $\bM\hbtheta=\hlambda\bN\hbtheta$, where $\bM$ is given in Eq. \eqref{M}. Here we will present general perturbation method to find general formulas for the variance and the bias. Therefore, we will keep the matrix $\bN$ unspecified, while in Section \ref{compare} we derive the bias for each method and compare between them.

We first start with the leading term of the variance. This term is also the leading term of the MSE as demonstrated in \cite{AliHo0, AliHo}. First we notice here that the leading term of the variance does not depend on $\bN$, while the leading term of the bias does. To accomplish our goals, we first expand $\hbtheta$ about the true values of the data. Using Taylor expansion, one can express
$
\hbtheta=\tilde{\btheta}+\Delta_1 \hbtheta+\Delta_2\hbtheta+\mathcal{O}_{ P}(\sigma^3),
$
where $\tilde{\btheta}$ is the true value of the parameters, while $\Delta_1 \hbtheta$ and $\Delta_2 \hbtheta$ are the first- and the second-order error terms of an estimator $\hbtheta$
Similarly, since $(\bM,\bN,\hlambda)$ are random variables (or matrices) of $\boldsymbol{x}_{ij}$'s, we expand
$\bM=\tbM+\Delta_1\bM+\Delta_2\bM+\cdots$, 
$\bN=\tbN+\Delta_1\bN+\Delta_2\bN+\cdots$
$\hlambda=\tlambda+\Delta_1\hlambda+\Delta_2\hlambda+\cdots.$
If we define $\bT=\Delta_2\bM-\Delta_1\bM\tbM^-\Delta_1\bM$, then  by using rigorous perturbation analysis \cite{Kanatani2011,AliHo0} to $\bM\hbtheta=\hlambda\bN\hbtheta$, one can show that the first- and the second-order error terms of an estimator $\hbtheta$ can be respectively expressed as
\begin{align}
\Delta_1\hbtheta&=\tbM^-\Delta_1\bM \tbtheta\label{Del1}\\
\Delta_2\hbtheta&= \minv\left(\tfrac{(\tilde{\bt}^\top\mathbf{T}\tilde{\bt})}{(\tilde{\bt}^\top\tilde{\bN}\tilde{\bt})} \tilde{\bN}\btth - \mathbf{T}\btth\right)\label{Del2}
\end{align}

The derivations of Eqs. \eqref{Del1} and \eqref{Del2} are lengthy but similar to those in \cite{Kanatani2011,AliHo}, and as such, their details were omitted here. Here $\tilde{\bM}$ is the noiseless version of the matrix $\bM$ and $\Delta_k\bM$ is the $k^{ th}$-order error term. Since $\tbxi_{ij}^\top \btth =0$, $\tbM\btth =\boldsymbol{0}$. Assuming that there are at least six distinct true points, the kernel of $\tbM$ is one dimensional, i.e., $ {\rm kernel }(\tbM)= {\rm span}(\btth )$. Therefore, the true matrix $\tbM$ is singular, and as such, the \emph{Moore-Penrose} inverse, abbreviated by $(^-)$, is used here. After lengthy derivations, the leading term of variance and the general form of the bias for the quadratic approximation of $\hbtheta$ are derived. Since $\Delta_1\bM$ is a linear combination of $\boldsymbol{e}_{ij}$'s, one obtains $\E[\Delta_1{\bt}] = -\minv\Delta_1{\bM}\E[\btth]=\mathbf{0}$. Using Eq. \eqref{Del1}, one can easily show that the leading term of the covariance matrix of the solution $\hat{\bt}$ is given by
\begin{equation}\label{Vtheta}
V[\hat{\bt}] = \E[\Delta_1{\hat{\bt}}\Delta_1{\hat{\bt}}^\top]=\sigma^2\tilde{\bM}^-\tilde{\bM'}\tilde{\bM}^- ,
\end{equation}
where $\tilde{\bM'} = \summ \left(\tilde{\bt}^\top \bV_0[\bxi_{ij}]\btth\right) \tbxi_{ij} \tbxi_{ij}^\top$. This shows that the leading term of the variance of $\hbtheta$ does not depend on $\bN$, so \emph{all algebraic methods minimizing $J$, with a quadratic constraint imposed on its parametric space, produce estimators that have the same leading term of the variance}. Next, we show that the bias depends on $\bN$, and as such, the bias of each method can be used as a tool to assess the accuracy of the fitting methods. Since $\E[\Delta_1{\bt}] $, the leading term of the bias is the so-called \emph{second-order} bias that takes the general expression
\begin{equation}\label{EDel2}
\mathbb{E}[\Delta_2\hbtheta]= \minv\left(\tfrac{\tilde{\bt}^\top\mathbb{E}[\mathbf{T}]\tilde{\bt}}{\tilde{\bt}^\top\tilde{\bN}\tilde{\bt}} \tilde{\bN}\btth - \mathbb{E}[\mathbf{T}]\btth\right).
\end{equation}
The derivation of the second-order bias $\E[\Delta_2{\hbtheta}]$ is quite lengthy and can be found in \cite{AliHo0,AliHo,KanataniSug2016b}. The second-order bias provides a general expression of the bias for any algebraic fit. Each algebraic fit has its own specific matrix $\bN$, and as such, each method has its own bias. In order to perform a comparison between methods, we derive an explicit equation for the expected value of $\mathbf{ T}$. Under the assumptions that the data $\boldsymbol{z}_{ij}$ is perturbed by independent and isotropic Gaussian noise, $\lim_{\sigma \to 0}\hat{\bt}(\boldsymbol{z}_{ij})={\btth}$ and $\btth^\top\bN\btth\neq 0$, then after lengthy calculations, one obtains
\begin{equation}\label{ETFinal}
\E[\mathbf{ T}] = \sigma^2\Bigl(\tilde{\bN}_{ S} - \boldsymbol{\Pi}-\summ \left({ \rm tr}[\minv \bV_0[\tbxi_{ij}]]\tbxi_{ij}\tbxi_{ij}^\top \right)\Bigr),
\end{equation}
where $\tilde{\bN}_{ S}=\tilde{\bN}_{ T} + 2S[\tbxi_c\boldsymbol{e}^\top]$, 
 $\tilde{\boldsymbol{\xi}}_c = \summ\tbxi_{ij}$, $\boldsymbol{e}=\bigl(1,0,1,0,0,0,0\bigr)^\top$, and
\begin{equation}\label{ETTerms}
\tilde{\boldsymbol{\Pi}}=\summ\tilde{ \boldsymbol{\Pi}}_{ij} ,\qquad
\tilde{\boldsymbol{\Pi}}_{ij}=\tbxi_{ij}^\top\minv\tbxi_{ij}\bV_0[\txi] + 2S\left[\bV_0[\tbxi_{ij}]\minv\tbxi_{ij}\tbxi_{ij}^\top\right].
\end{equation}
Here $S$ is the symmetrization operator, i.e. $S[A]=(A+A^\top)/2$). Now, since $\tbxi_{ij}^\top\btth = 0$,
\begin{equation}\label{ETtheta}
\E[\mathbf{ T}]\btth = \sigma^2\left(\tilde{\bN}_{ S}\btth - \tilde{\boldsymbol{\pi}} \right),
\end{equation}
where
\begin{equation}\label{pi}
\tilde{\boldsymbol{\pi}}= \tilde{\boldsymbol{\Pi}}\btth=\summ\left((\tbxi_{ij}^\top\minv\tbxi_{ij})\bV_0[\tbxi_{ij}]\btth+(\btth^\top\bV_0[\tbxi_{ij}]\minv\tbxi_{ij})\tbxi_{ij}\right).
\end{equation}
Moreover, after some mathematical manipulations and using the definition of $\tilde{\bM}'$, one can show that
\begin{equation}\label{TETO}
\btth^\top\E[\boldsymbol{T}]\btth = \sigma^2(\btth^\top \tilde{\bN}_{ T}\btth) - \sigma^2 {\rm tr}[\minv\tilde{\bM}'].
\end{equation}
The derivation of Eq. \eqref{TETO} is deferred to Appendix \ref{Apx.TETO}. 

\section{Comparisons between Methods }\label{compare}
With the aid of Eq. \eqref{ETtheta} and Eq. \eqref{TETO}, we are ready to present the second-order biases for each method. Since the second-order bias in Eq. \eqref{EDel2} depends on $\bN$, we devote this section to derive the biases of the LS, the O'Leary, and the Taubin methods. Then we compare their accuracies based on their biases. Many technical details are involved here and they are all deferred to Appendix \ref{BiasDerviation}.

\noindent \textbf{Bias of the Standard Least Squares Method.} The second-order bias of the LS fit is
\begin{equation}\label{LSBias}
\E[\boldsymbol{\Delta_2 {\hat{\theta}}}_{\rm LS}] = -\sigma^2\minv\left(\tilde{\bN}_{ \rm T}\btth + (\boldsymbol{e}^\top\btth)\boldsymbol{\tilde{\xi_{c}}} -\tilde{\boldsymbol{\pi}}\right)
\end{equation}
where $\tilde{\boldsymbol{\pi}}$ is defined in Eq. \eqref{pi}.\

\noindent \textbf{Bias of the O'Leary Method.}
The second-order bias of the O'Leary method is
\begin{equation}\label{OLearyBias}
 \E[\boldsymbol{\Delta_2{\hat{\theta}}}_{\rm OL}]= -\sigma^2\minv \left(p\tilde{\bN}_{\rm  OL}\btth  +(\boldsymbol{e}^\top\btth)\tilde{\boldsymbol{\xi}_c} - \tilde{\boldsymbol{\pi}}\right),
\end{equation}
where $\tilde{\boldsymbol{\pi}}$ is defined in Eq. \eqref{pi} and $ p=\frac{{\rm tr}[\minv\tilde{\bM}']}{\btth^\top\tilde{\bN}_{\rm OL}\btth}$.

It is worth mentioning here that $\tilde{\bM}^-\sim\mathcal{O}(\tfrac 1 n)$ and $\tilde{\bM}'\sim\mathcal{O}\left(n\right)$, thus $\tilde{\bM}^-\tilde{\bM}'\sim \mathcal{O}(1)$. Therefore $ {\rm tr}[\tilde{\bM}^-\tilde{\bM}']\sim\mathcal{O}(n)$. Finally, since $\bt^\top\bN_{ \rm OL}\bt\sim\mathcal{O}(1)$, then $p$ is of order $n$.  

\noindent{\textbf{Bias of the Taubin Method.}} Finally, we present the second-order bias of the Taubin method for concentric ellipses. i.e.,
\begin{equation}\label{TauBias}
\E[\boldsymbol{\Delta_2{\hat{\theta}}}_{\rm  T}] = -\sigma^2\minv\Bigl(q\tilde{\bN}_{ \rm T}\btth + (\boldsymbol{e}^\top\btth)\boldsymbol{\tilde{\xi_c}} -\tilde{\boldsymbol{\pi}}\Bigr),
\end{equation}
where $q=\frac{\rm  tr[\minv \boldsymbol{\tilde{\bM}'}]}{\btth^\top\tilde{\bN}_{\rm  T}\btth}$
and $\tilde{\boldsymbol{\pi}}$ is defined in Eq. \eqref{pi}. Furthermore, since ${\rm  tr}[\tilde{\bM}^-\tilde{\bM}']=\mathcal{O}(n)$. Since $\btth^\top\tbN_{\rm  T}\btth=\mathcal{O}(n)$, then $q$ is of order $1$.\bigskip

\noindent{ \bf Comparisons between Methods.} Taking a quick look at the biases of the three methods, one can easily see that they share some some commonalities and differences with each other. Table \ref{tab:table1} summarizes our findings. Before we compare between methods based on the different terms, we shall mention here that $ -\sigma^2(\boldsymbol{e}^\top\btth)\tilde{\bM}^- \tilde{\boldsymbol{\xi}_c} $ is of order of magnitude $\sigma^2$, while $\sigma^2\tilde{\bM}^- \boldsymbol{\pi}$ is of order $\mathcal{O}(\sigma^2/n)$. We classify terms into essential bias and nonessential biases as shown in Table \ref{tab:table1}. The essential bias is of order of magnitude $\sigma^2$, while the nonessential bias is of order $\sigma^2/n$. This means that the essential bias persists even if $n$ becomes large, while the nonessential bias diminishes for large $n$. Since all methods share the first leading term of the variance as well as many other terms in the bias, one can see in Table \ref{tab:table1} that the last column of Table \ref{tab:table1} distinguishes methods among each other.

\begin{table}[htb]
  \begin{center}
    \scalebox{0.77}{
    \begin{tabular}{|c|c|c|c|}
    \hline
      \multirow{2}{5em}{\textbf{Method}}&\multicolumn{2}{|c|}{\underline{\textbf{Common}}} & \multicolumn{1}{c|}{\underline{\textbf{Differences}}} \\ 
      & \textbf{  Essential Bias $\mathcal{O}(1)$  } & \textbf{  Non-Essential Bias
      $\mathcal{O}\left(\tfrac{1}{n}\right)$  } & \textbf{  Essential Bias $\mathcal{O}(1)$  } \\
     \hline
     \multicolumn{1}{|c|}{LS} & \multicolumn{1}{c|}{$-\sigma^2(\boldsymbol{e}^\top\btth)\tilde{\bM}^- \tilde{\boldsymbol{\xi}_c}$} & \multicolumn{1}{c|}{$\sigma^2\tilde{\bM}^- \tilde{\boldsymbol{\pi}}$}& \multicolumn{1}{c|}{ $-\sigma^2\tilde{\bM}^-\tilde{\bN}_{\rm T}\btth$} \\ \hline
      \multicolumn{1}{|c|}{OLE}& \multicolumn{1}{c|}{$-\sigma^2(\boldsymbol{e}^\top\btth)\tilde{\bM}^- \tilde{\boldsymbol{\xi}_c}$} & \multicolumn{1}{c|}{$\sigma^2\tilde{\bM}^- \tilde{\boldsymbol{\pi}}$}& \multicolumn{1}{c|}{ $-p\sigma^2\tilde{\bM}^-\tilde{\bN}_{{\rm OL}}\btth$}\\ \hline
      \multicolumn{1}{|c|}{TAU}&
     \multicolumn{1}{c|}{$-\sigma^2(\boldsymbol{e}^\top\btth)\tilde{\bM}^- \tilde{\boldsymbol{\xi}_c}$} & \multicolumn{1}{c|}{$\sigma^2\tilde{\bM}^- \tilde{\boldsymbol{\pi}}$}& \multicolumn{1}{c|}{$-q\sigma^2\tilde{\bM}^-\tilde{\tbN}_{\rm T}\btth$}\\ \hline
    \end{tabular}}
  \end{center} \vspace{-4mm}
   \caption{Anatomy of the biases.}
    \label{tab:table1}
    \vspace{-2mm}
\end{table}
Table \ref{tab:table1} shows that the essential bias of the LS and Taubin (TAU) methods differ in only one term. Since $q<1$ for large $n$, the bias of the Taubin method is smaller than the bias of the LS method. This shows why the Taubin method outperforms the LS method.

The comparisons between the O'Leary method and other methods are not straightforward. Therefore, we plotted the theoretical second-order biases, we can plot the biases for the three methods and we use our principle here;\emph{ the smaller the bias the better the performance. We compare methods in  the following three Scenarios:}\\

\begin{figure}[htb]\vspace{-4mm}
\begin{center}
\begin{tabular}{cc}
\hspace{-4mm}\includegraphics[width=.65\textwidth]{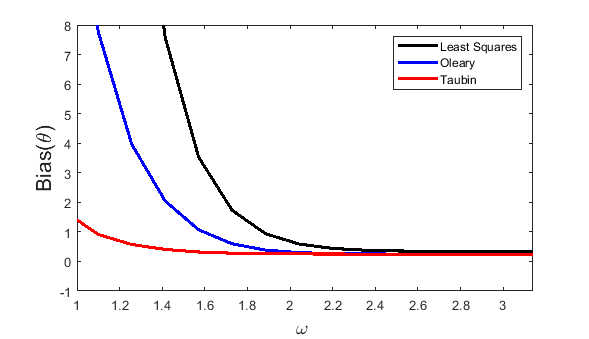} \hspace{-3mm}
\includegraphics[height=.35\textwidth, width=.38\textwidth]{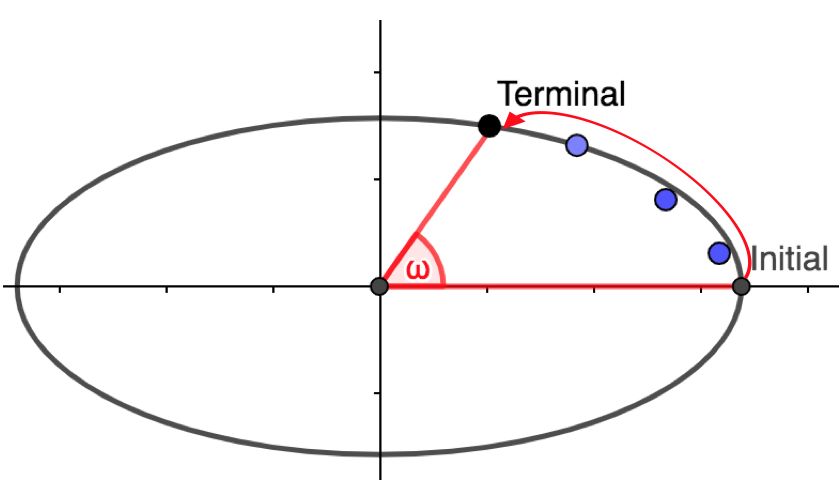}
\end{tabular}\vspace{-4mm}
\caption{(left)he theoretical second-order biases of the LS method, the O'Leary method, and the Taubin method as functions of the angle $\omega$ (in radians), where $\omega$ represents the terminal angle of the arcs that starts at $0$. We positioned $n_1=15$ and $n_2=20$ equidistant data points on two concentric ellipses centered at $(0,0)$ with $\tilde{a}_1=3$, $\tilde{a}_2=6$, $\tilde{b}_1=2$, and $\tilde{b}_2=4$.}
\label{fig:TBiasLTO} \end{center}
\end{figure}

\noindent\textbf{Scenario I.} We study the behavior of the bias for each method as a function of the arc length while keeping the model characterized by $\btth$ and $n_i$'s fixed. Figure \ref{fig:TBiasLTO} (right) reveals that the angle $\omega$ (in radians) represents the terminal angle of the arcs that starts at $0 $. In this scenario, Figure \ref{fig:TBiasLTO} (left) shows the theoretical bias of each method as a function of $\omega$. It is clear that the O'Leary method yields smaller bias than the LS method and the Taubin method yields the smallest bias of the algebraic methods. Moreover, as $\omega$ decreases, and as such, data are distributed over short arcs, the differences between methods become more obvious. \bigskip

\begin{figure}[t]\vspace{-4mm}\rb{
 \begin{tabular}{cc}
    \includegraphics[width=.5\textwidth]{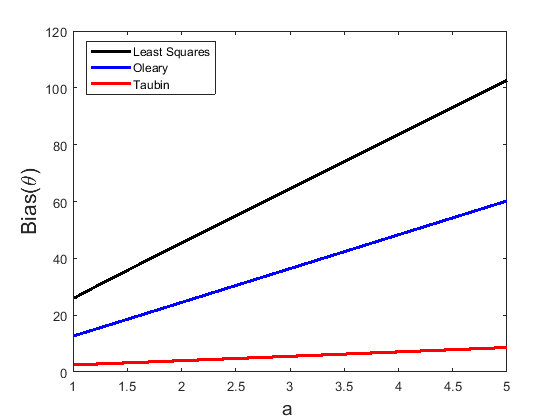} &\raisebox{5mm}{ \includegraphics[height=.27\textwidth, width=.5\textwidth]{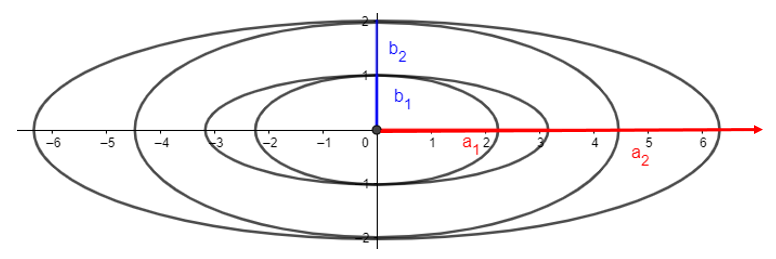} }
\end{tabular}\vspace{-3mm}
\caption{Theoretical second-order bias evaluated for different semi-major axes lengths $\tilde{a}_1\in[1.01,5]$ and $\tilde{a}_2=2\tilde{a}_1$, while the corresponding semi-minor axes $\tilde{b}_1=1$, $\tilde{b}_2=2$ are fixed with $n_1=15$ and $n_2=20$ equidistant data points distributed over an elliptic arc of $\frac{\pi}{2}$. The initial angle of the data starts at $0$ radians on ellipses centered at (0,0).} 
\label{fig:TBiaspi_2}}
\end{figure}

\noindent \textbf{Scenario II.} We study the behavior of the bias for each method as a function of semi-major axes $\tilde{a}_i$'s while fixing the other parameters and the sample sizes $n_i$'s fixed. The true points are distributed on the arcs located on the first quadrant of the Cartesian coordinating system. Without loss of generality, we set the semi-minor axis of the inner ellipse $\tilde{b}_1=1$. Figure \ref{fig:TBiaspi_2} (left) plots the theoretical second-order biases of the three methods as functions of $\tilde{a}_1\in[1,5]$. For simplicity, we assume the concentric ellipses are centered at $(0,0)$ and $\tilde{a}_2=2 \tilde{a}_1$, $\tilde{b}_2=2$, $n_1=15$, and $n_2=20$.

Figure \ref{fig:TBiaspi_2} (left) shows that the bias is a function of $a=a_1$ (see Figure \ref{fig:TBiaspi_2} (right).) The figure confirms that the Taubin method still yields the smallest bias. This agrees with our analytic results. It is worth mentioning here that we ran the same setting, but with different arcs' lengths, and we notice that all plots show the same pattern as in \ref{fig:TBiaspi_2} (left). Therefore, we only kept the case $\omega=\frac{\pi}{2}$ and omitted the other that result in the same conclusion regardless of the arcs lengths.    \bigskip

\begin{figure}[htb]
    \begin{tabular}{cc}\vspace{-4mm}
     \includegraphics[width=.55\textwidth]{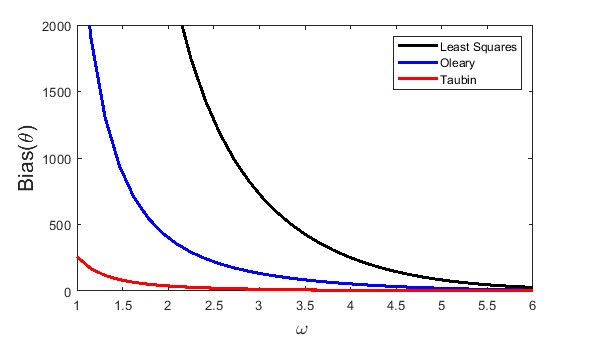}   \hspace{-2mm}\raisebox{4mm}{\includegraphics[height=0.24\textwidth, width=.45\textwidth]{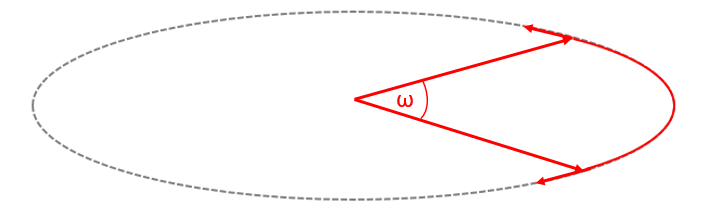} }
    \end{tabular}
\vspace{-2mm}    \caption{he theoretical second-order biases for high curvature.}
    \label{fig:Highcurve}
\end{figure}

\begin{figure}[htb]\vspace{-4mm}
    \begin{tabular}{cc}
     \includegraphics[width=.55\textwidth]{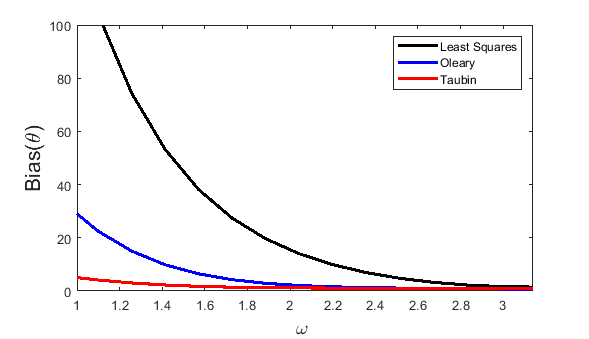}
 \hspace{-3mm}\raisebox{2mm}{     \includegraphics[height=0.28\textwidth, width=.45\textwidth]{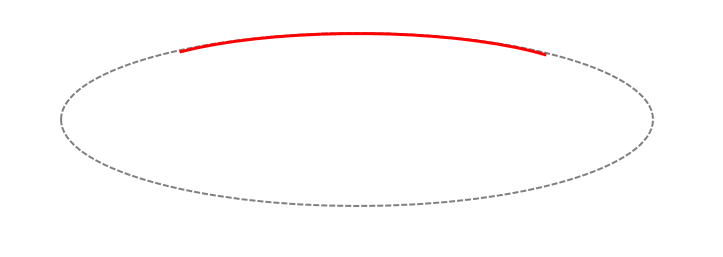} }
    \end{tabular}
   \vspace{-4mm} \caption{The theoretical second-order biases for low curvature.}
    \label{fig:Lowcurve}
\end{figure}

\noindent \textbf{Scenario III.} We study the behavior of the bias for each method as a function of the arcs' lengths $\omega$ while keeping the $n_i$'s fixed. Here we consider two cases (i) the data are distributed over high curvature (see Figure \ref{fig:Highcurve} (right)) and (ii) low curvature arcs (see Figure \ref{fig:Lowcurve} (right).) Here $n_1=15$ and $n_2=20$ equidistant data points were positioned on the inner and outer ellipses, respectively. The concentric ellipses share the same center $(0,0)$, while we set $\tilde{a}_1=3$, $\tilde{b}_1=1$, $\tilde{a}_2=6$, and $\tilde{b}_2=2$. Figures \ref{fig:Highcurve} (left) and \ref{fig:Lowcurve} (left) show the behavior of the bias for each method in the  high and low curvature cases. The two figures the biases for the three methods follow the same pattern as we have seen in Scenarios I-II. However, the magnitudes of the biases in the high curvature case are much larger than those in the low curvature case.

\section{Hyper-accurate Methods}\label{HYPER}
Our theoretical error analysis indicates that the Taubin method is superior to the LS and O'Leary methods and yields the most accurate fit. However, we extend our analysis even further to find an even more accurate fitting method. Now, we utilize the results of our error analysis to develop a new normalization that produces an estimator with zero second-order bias. We refer to this method as the Hyper-accurate Least Squares fit (or hyper for short) for concentric ellipses. Unlike the other methods, the Hyper LS method does not minimize any objective function. Its solution comes as the result of solving the generalized eigenvalue problem in Eq. \eqref{singleGEP} with the matrix $\bN$ that will be chosen carefully here. The new method eliminates the entire bias of order $\sigma^2$ and $\sigma^2/n$. Because of its complexity, we also propose a simplified version of Hyper LS that produces virtually the same results for relatively large sample sizes. Following \cite{Kanatani2011,KanataniAli}, we call this method the Semi-Hyper method. \medskip

\subsection{Hyper Least Squares}
It is interesting to see if we can choose $\bN$ such that the second-order bias is zero. Taking a careful look at the general bias given in Eq. \eqref{EDel2}, we can see that if we choose
$\bN_{\rm H}$ so that its noiseless value $\tilde{\bN}_{ H}$ satisfies $\E[\boldsymbol{ T}\btth] =\sigma^2 \tilde{\bN}_{\rm  H}\btth$, then $\E[\boldsymbol{\Delta_2^{\perp}{\theta}}]=\mathbf{0}$.
Therefore, our normalization matrix is
\begin{equation}\label{NH}
\tilde{\bN}_{\rm  H} = \tilde{\bN}_{ \rm T} + 2S[\boldsymbol{\tilde{\xi}_ce^\top}] - \summ \bigl( {\rm tr}[\minv \bV_0[\txi]\txi\txi^\top + \tilde{\boldsymbol{\Pi}}_{ij}\bigr).
\end{equation}
Multiplying Eq. \eqref{NH} by $\btth$ yields
$$\tilde{\bN}_{ \rm H}\btth =  \Bigl(\tilde{\bN}_{ \rm T}\btth + 2S[\boldsymbol{\tilde{\xi}_ce^\top}]\btth - \summ \left({\rm tr} [\minv \bV_0[\txi]\txi\txi^\top\btth + \tilde{\boldsymbol{\Pi}}_{ij}\btth\right)\Bigr).
$$Since $\txi^\top\btth=0$, we have
$\tilde{\bN}_{\rm H}\btth=\tilde{\bN}_{\rm T}\btth+(\boldsymbol{e}^\top\btth)\tilde{\boldsymbol{\xi}_c} - \tilde{\boldsymbol{\pi}}.$ Therefore, recalling Eq. \eqref{ETtheta} we can see $\sigma^2\tilde{\bN}_{\rm H}\btth=\E[\boldsymbol{ T}\btth]$. Then, the bias is as follows
\begin{equation}\label{HyperBias}
\E[\boldsymbol{\Delta_2{\theta}}] = -\sigma^2 \minv \left(\frac{\btth^\top\tilde{\bN}_{\rm H}\btth}{\boldsymbol{\tilde{\theta}}^\top \tilde{\bN}_{\rm  H}\btth}\tilde{\bN}_{\rm H}\btth - \tilde{\bN}_{\rm H}\btth \right) = \mathbf{0}.
\end{equation}
Thus, the solution obtained with the use of the noisy version of $\tilde{\bN}_{ H}$ returns estimators with zero second-order bias. \\

\noindent\textbf{Implementation of the Hyper LS Method.} Unlike other algebraic methods, the Hyper LS method does not minimize any objective function. Instead, we need to solve the generalized eigenvalue problem $\bM\bt=\lambda\bN_{\rm H}\bt$, where $\bN=\bN_{\rm H}$. Since our general analysis shows that the true value of $\lambda$ must be zero, it is more natural to select the generalized eigenvalue that is closest to zero regardless of its sign. That is, our solution is the generalized eigenvector corresponding to the generalized eigenvalue closest to zero. The matrix $\bN_{\rm  H}$ is indefinite and has a complex structure. Since $\bM$ is positive definite, a remedial resolution to this problem is to rewrite the generalized eigenvalue problem as $ \bN_{\rm H}\bt=\eta\bM\bt$ with $\eta=\tfrac{1}{\lambda}$ and find the generalized eigen vector corresponding to the largest absolute generalized eigenvalue. Moreover, it is worth mentioning here that in the practical implementation of the Hyper LS method, we compute the generalized inverse $\bM^-$ even though it is a noisy version of $\tilde{\boldsymbol{M}}$, and as such, it is theoretically non-singular. The rationale of this is that in practice $\boldsymbol{M}$ has a large conditioning number and it is nearly singular. Using the eigenvalue decomposition, we compute $\boldsymbol{M}=\bU\bD\bU^\top$, where $\mathbf{D}={\rm Diag}(\beta_1,\ldots,\beta_7)$. If the smallest eigenvalue $\beta_7<10^{-6}$, we set $\bD^{-}={\rm Diag}\left(\tfrac 1{\beta_1},\ldots,\tfrac{1}{\beta_6},0\right)$. Then, the truncated generalized inverse $\bM^-=\bU\bD^{-} \bU^\top$ will be used.\

\subsection{Semi-Hyper Least Squares}
Apparently, the Hyper method uses a more complex normalization matrix, and as such, it is expected to be computationally more expensive than the others especially when the sample sizes are relatively large. In this case, we can now define a new constraint that is the large sample version of $\bN_{\rm H}$. That is, when $n$ is relatively large, its nonessential bias $\mathcal{O}(\sigma^2/n)$ becomes negligible. After careful investigation, one can show that solving for the generalized eigenvector in $\bM\bt=\lambda\bN_{\rm  S}\bt$ with the following constraint matrix produces an estimator with zero essential bias.
\noindent We call the method using constraint $\bN_{\rm S}$ the Semi-Hyper LS method.
\begin{equation}\label{NS}
\bN_{\rm S} = \bN_{\rm T}+2S[\tilde{\boldsymbol{\xi}}_c\boldsymbol{e}^\top].
\end{equation}

\noindent\textbf{Implementation of the Hyper LS Method.} The numerical implementation for the Semi-Hyper LS follows the same footsteps of the Hyper LS method implementation.

Now after some analysis one can show that the second-order bias of the Semi-Hyper LS method is
\begin{equation}\label{SemiHyperBias}
E[\boldsymbol{\Delta_2{\hat{\theta}_{S}}}]= -\sigma^2\minv \left(q\tilde{\bN}_{ \rm T}\btth +(\boldsymbol{e}^\top\btth)\boldsymbol{\tilde{\xi}}_c)  -  \tilde{\boldsymbol{\pi}}\right).
\end{equation}
(Its derivation is deferred to the appendix.) Combining our results together, we summarize the second-order bias of the five methods and their decomposition in Table \ref{tab:table2}. 
 \begin{table}[H]
  \begin{center}
    \scalebox{.79}{
    \begin{tabular}{|c|c|c|c|}
    \hline
      \multirow{2}{5em}{\textbf{Method}}&\multicolumn{2}{|c|}{\underline{\textbf{Common}}} & \multicolumn{1}{c|}{\underline{\textbf{Differences}}} \\ 
      & \textbf{Essential Bias $\mathcal{O}(1)$} & \textbf{Non-Essential Bias
      $\mathcal{O}(\tfrac{1}{n})$} & \textbf{Essential Bias $\mathcal{O}(1)$} \\
     \hline
     \multicolumn{1}{|c|}{LS} & \multicolumn{1}{c|}{$-\sigma^2(\boldsymbol{e}^\top\btth)\tilde{\bM}^-\tilde{\boldsymbol{\xi}_c}$} & \multicolumn{1}{c|}{$\sigma^2\tilde{\bM}^-\tilde{\boldsymbol{\pi}}$}& \multicolumn{1}{c|}{ $-\sigma^2\tilde{\bM}^-\tilde{\bN}_{ \rm T}\bt$}\\ \hline
      \multicolumn{1}{|c|}{OLE}& \multicolumn{1}{c|}{$-\sigma^2(\boldsymbol{e}^\top\btth)\tilde{\bM}^-\tilde{\boldsymbol{\xi}_c}$} & \multicolumn{1}{c|}{$\sigma^2\tilde{\bM}^-\tilde{\boldsymbol{\pi}}$}& \multicolumn{1}{c|}{ $-p\sigma^2\tilde{\bM}^-\tilde{\bN}_{{\rm OL}}\bt$}\\ \hline
     \multicolumn{1}{|c|}{TAU}&
     \multicolumn{1}{c|}{$-\sigma^2(\boldsymbol{e}^\top\btth)\tilde{\bM}^-\tilde{\boldsymbol{\xi}_c}$} & \multicolumn{1}{c|}{$\sigma^2\tilde{\bM}^-\tilde{\boldsymbol{\pi}}$}& \multicolumn{1}{c|}{$-q\sigma^2\tilde{\bM}^-\tilde{\bN}_{ \rm T}\bt$}\\ \hline
     \multicolumn{1}{|c|}{SEMI}&
     \multicolumn{1}{c|}{$-q\sigma^2(\boldsymbol{e}^\top\btth)\tilde{\bM}^-\tilde{\boldsymbol{\xi}_c}$} & \multicolumn{1}{c|}{$\sigma^2\tilde{\bM}^-\tilde{\boldsymbol{\pi}}$}& \multicolumn{1}{c|}{$-q\sigma^2\tilde{\bM}^-\tilde{\bN}_{\rm  T}\bt$}\\ \hline
   \multicolumn{1}{|c|}{HYPER}&
     \multicolumn{1}{c|}{$\boldsymbol{0}$}& \multicolumn{1}{c|}{$\boldsymbol{0}$}& \multicolumn{1}{c|}{$\boldsymbol{0}$}\\ \hline
    \end{tabular}}
  \end{center}\vspace{-5mm}
      \caption{Anatomy of the biases for the five methods.} \label{tab:table2}
\end{table}

\begin{figure}[htb]
\includegraphics[height=6cm,width=.9\textwidth]{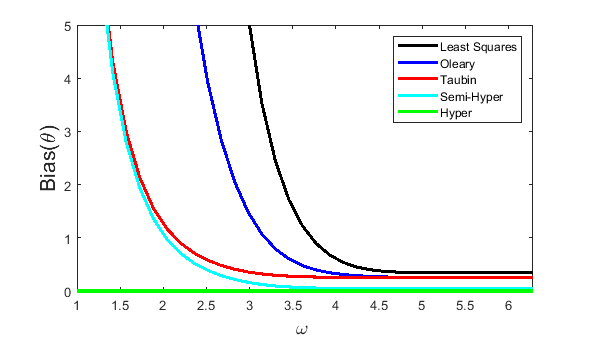}
\vspace{-5mm}\caption{Theoretical Bias for each algebraic method versus elliptic arc lengths $\omega$.} \label{BiasF}
\end{figure}

To understand how the theoretical bias behaves for the five methods, we plotted in Figure \ref{BiasF} the bias of the five methods as a function of the arc length\footnote{We follow the same setting of Scenario I. The other scenarios led to similar conclusions, and as such, they are omitted in this paper.}. The figure demonstrates that the Taubin and the Semi-Hyper LS methods produce very similar results for short arc length but when the arc length becomes larger than $\tfrac \pi 2$, the Semi-Hyper LS yields smaller bias. Last but not least, the Hyper LS method outperforms all other methods and yields an unbiased estimator up to the second order. 

\section{Numerical Experiments}\label{NumeExpeConc}
We now turn our attention to validating our theoretical results. All algebraic methods will be assessed by a series of numerical experiments. We applied the five methods on real data extracted from the image of a human iris. We will then compare between methods using Monte Carlo simulations.

\subsection{Numerical Experiments on Real Data}
\begin{figure}[htb]
\begin{center} 
\includegraphics[height=2.6in,width=4in]{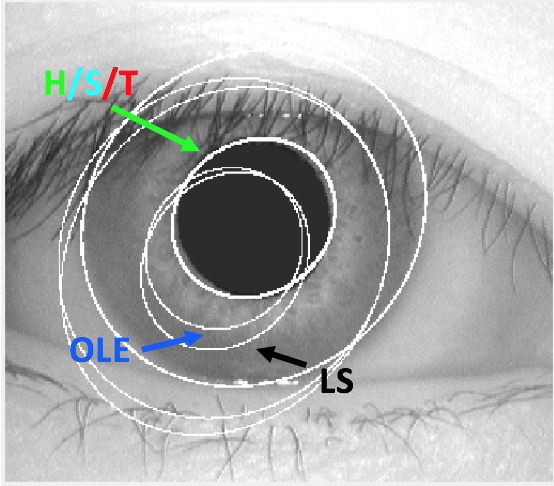} \vspace{-3mm}
\caption{Superimposed image of all methods on inner boundary for two image of human's irises. The Hyper LS (H), Semi-Hyper LS (S), the Taubin (T), the O'Leary (OL), and LS methods (LS).} \label{fig:Iris2a}
\end{center}
\end{figure}

\begin{figure}[htb]
\begin{center}\vspace{-1in}
\hspace{-0.75in}\includegraphics[height=4in,width=7in]{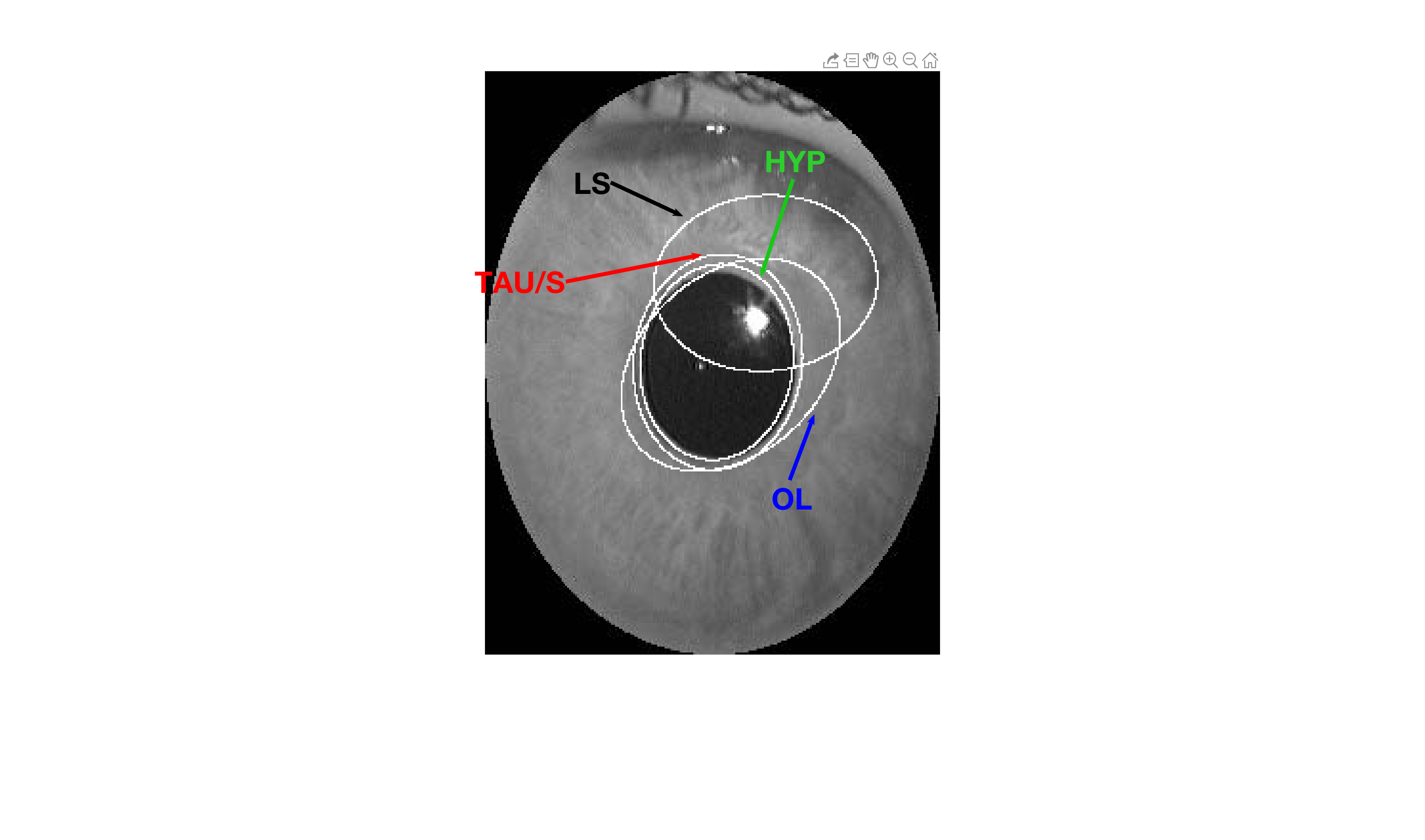} \vspace{-2.5cm}
\caption{\ Superimposed image of all methods on inner boundary for a human's iris. The Hyper LS (H), Semi-Hyper LS (S), the Taubin (T), the O'Leary (OL), and LS methods (LS).} \label{fig:Iris2b}\end{center}
\end{figure}

To corroborate our analytic findings with a set of practical data points, we applied all algebraic methods on a real images and we assessed their performances. Since one of the most important applications for fitting concentric ellipses appears in iris recognition \cite{CASIA,HarkerWeb,Kong12,Pillai11}, we applied all methods on the binary image of two human's irises. The images were captured by a digital camera with a $320\times 290$ resolution. The \emph{Canny Edge Detection} technique \cite{Canny86} was implemented to extract the digitized data points from the inner and outer concentric ellipses. We set $f_0=100$ as the scale factor so that   the data set is normalized \footnote{This image resolution of the images was $320\times 290$ pixels, so we used the scale factor $f_0=100$ pixels to normalize the extracted data, and as such, the transformed image stays within a reasonable numerical range.}. We have a total of $n_1 = 100$ and $n_2 = 150$ points from both concentric ellipses, respectively. The clean image is corrupted by white noise with a zero mean. Here, the noise level $\sigma$ was set 10 pixels. The 1000 ensemble runs were performed, then the average of each method was computed. As Figures \ref{fig:Iris2a} and \ref{fig:Iris2b} show, we superimposed the five methods on each image: the Hyper LS (H), the Semi-Hyper LS (S), the Taubin (T), the O'Leary (OL), and the LS methods (LS).
In both images, one can observe that the LS and the O'Leary methods produce estimates with heavy biases. Both methods are far away from the actual inner and outer boundaries. This is not an unexpected observation regarding the LS method which is widely known to be heavily biased. The O'Leary method performs better than the LS method, however, it returns a biased result. Figure \ref{fig:Iris2a} shows that the Hyper LS, the Semi-Hyper LS, and the Taubin methods all produce similar results. These methods produce results that are very close to the inner and outer boundaries of the image of the human iris. The results in Figure \ref{fig:Iris2b} areslightly different, where the Hyper LS method provides the closest fit to the inner ring of the iris, followed by the Taubin and Semi-Hyper LS methods.

\subsection{Numerical Experiments on Synthetic Data}
\begin{figure}[H]\vspace{-2mm}
    \begin{tabular}{ccc}
      \includegraphics[height=.27\textheight,width=.33\textwidth]{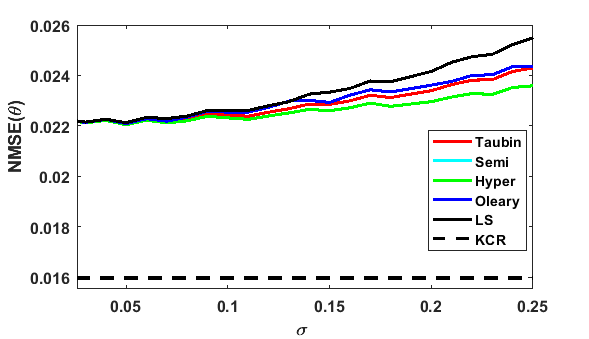}    & \includegraphics[height=.27\textheight, width=.33\textwidth]{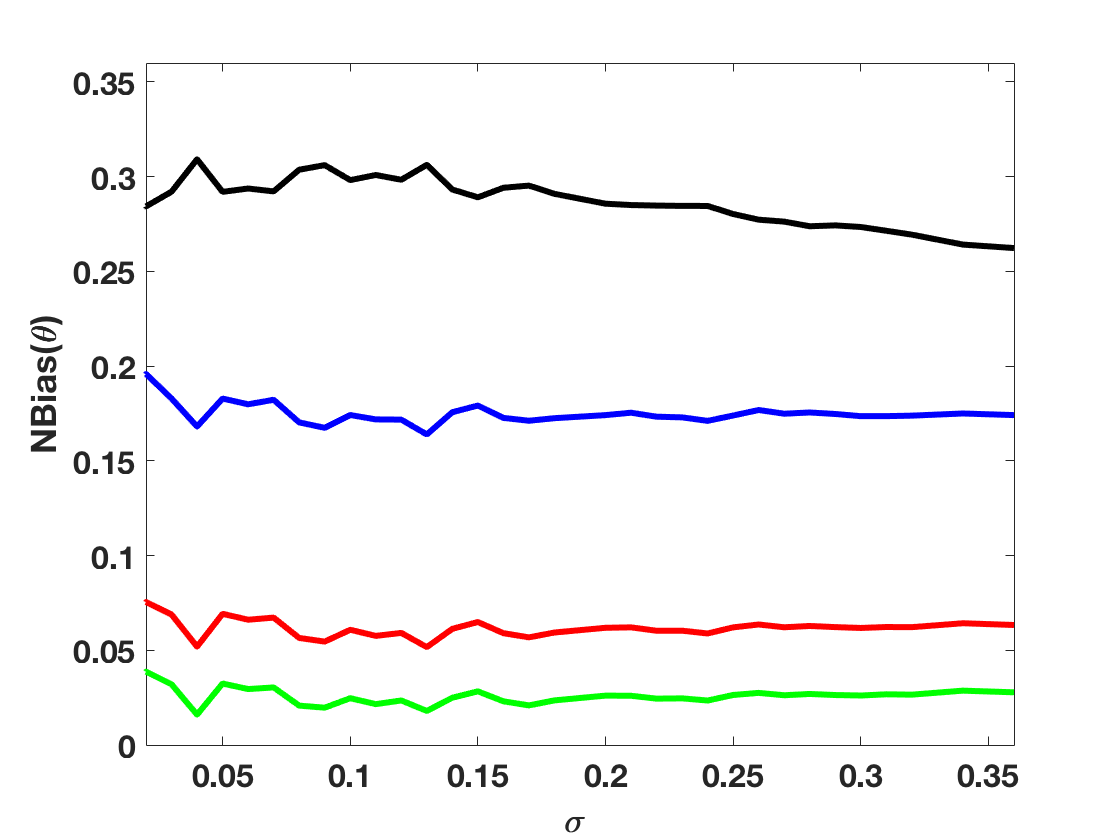}&\includegraphics[width=0.33\textwidth, height=.27\textheight]{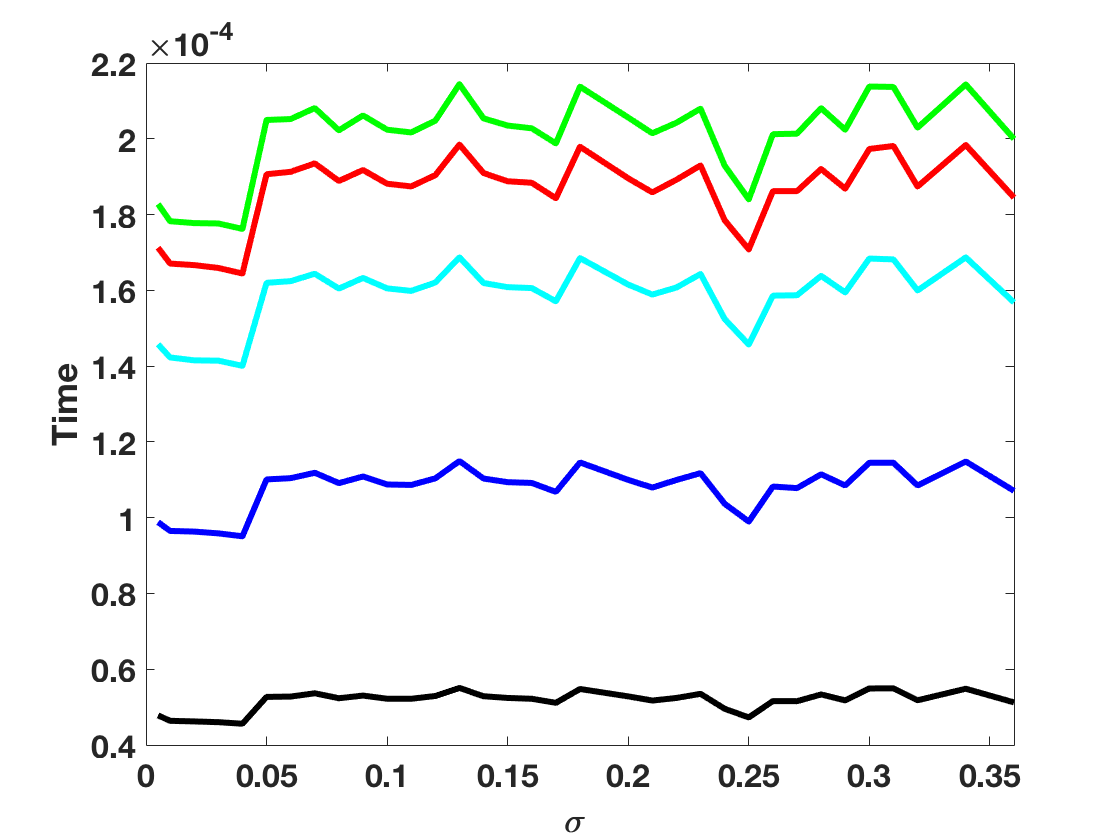}
    \end{tabular}
\vspace{-4mm}\caption{The NMSE($\hat{\bt}$) (left) and the${\rm NB}(\hat{\bt})$ (middle) for all methods and ART (right)for all methods in used Experiment I.}
    \label{fig:NMSE1}
\end{figure}

\begin{figure}[htb]
    \centering
    \begin{tabular}{ccc}
    \includegraphics[height=.27\textheight,width=.33\textwidth]{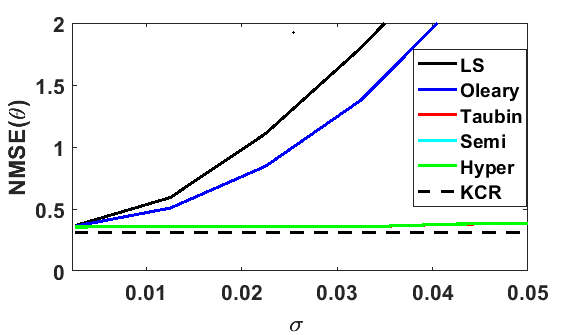} &  \includegraphics[height=.27\textheight,width=.33\textwidth]{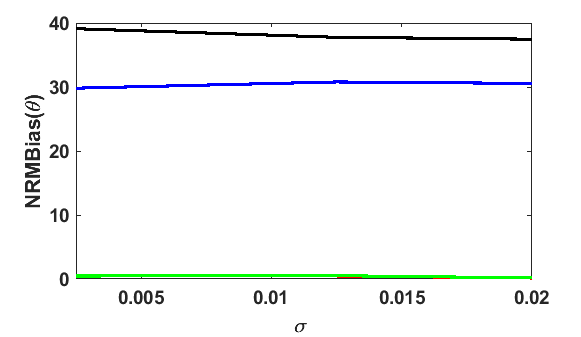}&{\vspace{-5mm}\includegraphics[height=.27\textheight, width=0.33\textwidth]{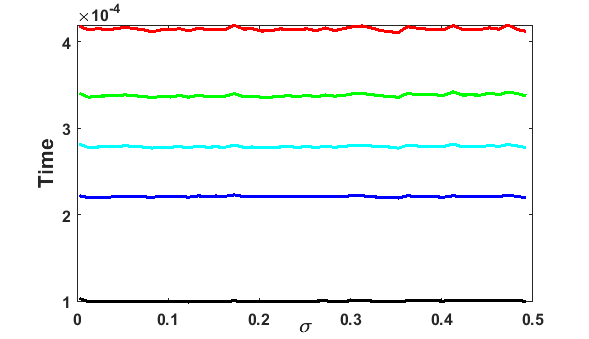}}
    \end{tabular}
   \vspace{-3mm} \caption{The NMSE($\hat{\bt}$) (left) and the $\rm NB(\hat{\bt})$ (middle) and ART (right) for all methods used in Experiment II.}
    \label{fig:NMSE2}
\end{figure}

The experimental results from the real data confirm our theoretical study but it does not reveal the whole picture. To get more insight, we ran more numerical experiments using Monte Carlo simulations. We ran two experiments and we
computed the \emph{Normalized Mean-Squared Error} (NMSE) and the \emph{Normalized Bias (NB)}. i.e.,
    \begin{equation*}\label{NMSE}
    \rm NMSE(\hat{\bt})= \frac{1}{\sigma^2N}\sum_{b=1}^N \left\|\hat{\bt}^{(b)}- \btth\right\|^2, \qquad  \rm NB(\hat{\bt}) = \left\|\frac{1}{\sigma^2N}\sum_{b=1}^N 
    \hat{\bt}^{(b)}-\btth\right\|.
\end{equation*}
We also computed the \emph{Average Run-Time} (ART) and the \emph{Convergence Rate} (i.e., percentage of returning concentric ellipses in the $N$ fitted geometric shapes). In the two experiments, we positioned $n_1$ and $n_2$ equidistant true points over elliptic arc of length $\omega$ starting at the initial angle 0. From these points, we generated data points $\left\{(x_{ij},y_{ij})\right\}_{j=1}^{n_i}$, for $i=1,2$, by adding to $(x,y)$ white noise with a zero mean  and a standard deviation $\sigma$. For each value of $\sigma$, $N=10^5$ ensemble runs were performed and for each run, the estimates of the parameters were computed for each method. Figures \ref{fig:NMSE1} (left) and Figure \ref{fig:NMSE2} (left) represent the NMSEs of the five methods, while Figures \ref{fig:NMSE1} (middle) and Figure \ref{fig:NMSE2} (middle) represent their normalized biases NBs. In all experiments, the horizontal axes represent the noise level $\sigma$. The dashed line represents the variance lower bound computed according to \cite{Chernov2}. Moreover, Figures \ref{fig:NMSE1} (right) and Figure \ref{fig:NMSE2} (right) show the ART for each method.\\

\noindent\textbf{Experiment I: Monte Carlo simulation for long arcs.} In this experiment, we positioned $n_1=10$ and $n_2=15$ true points on long elliptical arcs with $\omega=\frac{5\pi}{3}^\circ$. The true concentric ellipses are centered at $(-3,3)$ with semi-major axes $\tilde{a}_1=5$ and $\tilde{a}_2=10$, while the semi-minor axes $\tilde{b}_1=1$ and $\tilde{b}_2=2$. Figure \ref{fig:NMSE1} (left) shows that the NMSEs for all methods. Since the MSE for all methods have a common leading term, which is of order $\sigma^2$, its normalization is of order 1. The leading term of the NMSE is ${\rm tr}[\bV_0[\hbtheta]]$, which is also the leading term of the variance. Therefore, as the noise level converges to zero, the first leading term of their NMSEs converges to the normalized leading term of the variance, which is a common term in the MSE for each method. Accordingly, methods start behaving differently when the noise level $\sigma$ increases. This is due to the magnitude of the bias in each method. As Figure \ref{fig:NMSE1} (middle) reveals, methods differ in their biases, i.e.,
$
{\rm NB}(\hbtheta_{\rm H}) ={\rm NB}(\hbtheta_{\rm S})<{\rm NB}(\hbtheta_{\rm T})<{\rm NB}(\hbtheta_{\rm OL})<{\rm NB}(\hbtheta_{\rm LS})
$
That is, as the noise level $\sigma$ increases, the effect of bias becomes the major factor in the MSE's behavior, and as such, the corresponding  MSEs are ranked accordingly (see Figure \ref{fig:NMSE1} (left).) It shows that the NMSE for the Hyper LS and the Semi-Hyper LS methods coincide each other and they have the smallest NMSE methods. The Taubin method produces a less accuracy than the Semi-Hyper and Hyper LS methods, but still is more accurate than the O'Leary method. The LS method, on the other hand, provides the least accurate fit and has the highest NMSE. Moreover, Figure \ref{fig:NMSE1} depicts the average run-time for each method. Clearly, the LS method is by far the fastest method, followed by The O'Leary method, the Semi-Hyper method, the Hyper method, and the Taubin method (in this order).

%
\noindent\textbf{Experiment II: Monte-Carlo simulation for short arcs.} In this experiment, we positioned $n_1=15$ and $n_2=20$ true points on short elliptical arcs with $\omega=\frac{\pi}{2}^{\circ}$. The true concentric ellipses are centered at $(0,0)$, with semi-major axes $\tilde{a}_1=3$ and $\tilde{a}_2=6$, while the semi-minor axes $\tilde{b}_1=2$ and $\tilde{b}_2=4$. Figure \ref{fig:NMSE2} (left) shows that the Hyper LS, the Semi-Hyper LS, and the Taubin methods produce very close results yielding the highest accuracy. The O'Leary method is less accurate than this pack but it still produces better results than the least accurate one, i.e., the LS. Furthermore, Figure \ref{fig:NMSE2} (right) shows the average run time for each method. As can be seen from the figure, the LS method yields the fastest results, followed by the O'Leary method, the Semi-Hyper method, the Taubin method, and the Hyper LS method (in this order).
%
Before we conclude this section, two important remarks shall be addressed here:\\

\noindent\textbf{Remark I: Divergence Issue.}
First, recall that each constraint matrix in these methods is constructed to yield a well-defined solution. However, all methods, except the O'Leary method, do not guarantee producing concentric ellipses. That is, the quadratic constraints were designed to eliminate the indeterminacy in the solution (as in the LS and Taubin methods) or to eliminate the bias (as in the Hyper and Semi-Hyper LS methods). Therefore, in both numerical experiments, we computed the percentage of times each methods returned concentric ellipses. Our results show that all methods were always returning concentric ellipses for the long arc experiments but the situation for short arcs was different. 

Table \ref{tab:ConvRate} presents the relative frequency of returning concentric ellipses for each method in the short-arcs experiment. One can obviously see here that the O'Leary method produces concentric ellipses 100\% of the times regardless of the noise level. The other methods, on the other hand, are not necessarily returning concentric ellipses. The rate of convergences for other methods depend on the noise level. Table \ref{tab:ConvRate} represents the convergence rate for all methods in experiment II. The table shows that as the noise level $\sigma$ increases, the LS method fails drastically. The Hyper LS, Semi-Hyper and Taubin methods are also more prone to divergence whenever a data set is observed with large levels of noise. However, their sensitivity to returning non-elliptical shapes was much less than the LS. That is, while the convergence rate for the LS was 2\% when $\sigma=0.3$, the three methods (TAU, Semi, Hyper) are still producing concentric ellipses 75\% of the time. 

\begin{table}[H]
\begin{center}
\scalebox{.9}{
\begin{tabular}{ |c||c|c|c|c|c| }
 \hline
$\boldsymbol{\sigma}$& \textbf{LS}&	\textbf{OLE}&\textbf{TAU} &\textbf{Semi}&\textbf{Hyper}  \\
\hline \hline
0.005	&100 &100	&100	&100	&100  \\
\hline
0.1	&48	&100	&99 &99	&99  \\
\hline
0.2	&4	&100	&90	&90	&90  \\
\hline
0.3	 &2	&100	&75	&76	&75 \\
 \hline
\end{tabular}}
\end{center}\vspace{-6mm}
 \caption{Convergence rate to return concentric ellipses (in \%).}   \label{tab:ConvRate}
\end{table}


\noindent\textbf{Remark II: The Case $K\geq 3$.} In many other real-life applications, objects might appear as three or more concentric ellipses. This paper sheds more lights on the problem of fitting two concentric ellipses. The generalization of our theory here to $K\geq 3$ is straightforward. Firstly, the algebraic parameter vector is $\bt=(A,B,C,D,E,F_1,\ldots,F_K)^\top$, the mathematical representations of expressions  the same but the only difference is $\bM_{ij}$:
$
\bM_{1j}^{22} = {\rm Diag}\bigl(f_0^4,0,\ldots,0\bigr) ,  \,\ \bM_{2j}^{22} ={\rm  Diag}\bigl(0,f_0^4,0,\ldots,0\bigr),\,\bM_{Kj}^{22} ={\rm  Diag}\bigl(0,\ldots,f_0^4\bigr).
$
The practical implementations for all methods will be similar to those developed in this paper.

\section{Discussions and Conclusions}\label{Conclusion}
Iris detection is one of the most important steps in iris recognition systems, which play a vital role in the biometric recognition system for the US security and robotic industries. There are some methods for this object identification problem, however, they perform poorly when images are blurry. Since this problem lies under the umbrella of the problem of geometric estimation, addressing this problem turns out to involve  fitting concentric ellipses to data. We adopted some statistical assumptions tailored to those applications. In order to get very accurate results, some iterative methods can be used to get the highest accuracy. Such methods are the maximum likelihood estimator, the gradient weighted algebraic methods, or the Renormalization method. However, these methods are iterative and their convergences depend heavily on the initial guess they are seeded. They are also prone to divergence when the noise level is moderate. Therefore, this paper is devoted to studying the algebraic non-iterative methods.

In this paper, the existing methods were reviewed and the well-known Taubin method was extended to our problem. Then, according to perturbation theory, our error analysis was performed and two non-iterative methods were developed. The practical implementations of all methods were discussed. We compared between methods theoretically and investigated their performances under different circumstances. Those new methods can be used as reliable initial guesses for other iterative methods, and as such, their convergence rates will be enhanced. However, when images are corrupted by relatively large noise, iterative methods fail, then other trustworthy approaches are sought. We show in this paper that our methods developed here do not only outperform all existing ones, but they are also robust against large noise. Our methods and their practical performances are assessed by a series of numerical experiments on both synthetic and real data.

\section{Appendix}\label{Apx}

\subsection{Relationship Between Algebraic and Geometric Parameters of an ellipse}\label{relation}
It is important to understand the relationship between the geometric parameters and algebraic parameters. When performing a single ellipse fit using the algebraic parameters $\bt=(A,B,C,D,E,F)^\top$, we plot the ellipse by its geometric parameters which helps us understand the geometric structure of the ellipse. The equations below show the relationship between the spaces parameterized by the algebraic parameters $\bt$ and the geometric parameters $\boldsymbol{\phi} = (x_c,y_c,a,b,\psi)^\top$. The detail of conversion from the algebraic to the geometric parameters can be found in \cite{Kanatani2011}.
\begin{equation}\label{AtoGConv}
\begin{split}
&x_c = \tfrac{CD - BE}{B^2-AC} \\
&y_c = \tfrac{AE - BD}{B^2-AC} \\
&a = -\tfrac{\sqrt{2(AE^2+CD^2-BDE+(B^2-AC)F)(A+C+\sqrt{(A-C)^2+B^2})}}{B^2-AC} \\
&b = -\tfrac{\sqrt{2(AE^2+CD^2-BDE+(B^2-AC)F)(A+C-\sqrt{(A-C)^2+B^2})}}{B^2-AC} \\
&\psi = \arctan{\tfrac{C-A-\sqrt{(A-C)^2+B^2}}{B}}, \ \rm{for \ B} \neq 0.
\end{split}
\end{equation}

\noindent\textbf{Geometric to Algebraic Parameter Conversion.} \label{appendix:GtoA}
The conversion formulas from geometric parameters to algebraic parameters are straightforward.
  \begin{equation}
  \begin{split}\label{GtoAConv}
   A &= \tfrac{\cos^2(\psi)}{a^2} + \tfrac{\sin^2(\psi)}{b^2},\qquad \qquad 
   B = \cos(\psi)\sin(\psi)\bigl(\tfrac{1}{a^2} - \tfrac{1}{b^2}\bigr) \\
  C &=\tfrac{\sin^2(\psi)}{a^2} + \tfrac{\cos^2(\psi)}{b^2} ,\qquad \qquad 
  D = -Ax_{c}-By_{c}\\
   E &= -Cy_{c} - Bx_{c}  ,\qquad \qquad \quad
   F =Ax_{c}^2+Cy_{c}^2+2Bx_{c}y_{c}-1 .
\end{split}
\end{equation}
This conversion from the algebraic parameter space to the geometric parameter space can be justified as follows. Let us first denote $c = \cos(\psi)$ and $s = \sin(\psi)$ and expand the following equation:
\begin{align*}
1&=\tfrac{\left[(x-x_{c})\cos(\psi) + (y-y_{c})\sin(\psi)\right]^2}{a^2}+\tfrac{\left[-(x-x_{c})\sin(\psi) + (y-y_{c})\cos(\psi)\right]^2}{b^2}\\
0&=\tfrac{(b^2c^2+a^2s^2)}{a^2b^2}x^2+\tfrac{2(b^2cs-a^2cs)}{a^2b^2}xy +\tfrac{(b^2s^2+a^2c^2)}{a^2b^2}y^2 - \tfrac{2(b^2c^2x_c+a^2s^2x_c-b^2csy_c-a^2csy_c)}{a^2b^2}x\\
&-\tfrac{2(b^2csx_c-a^2csx_c+b^2s^2y_c+a^2c^2y_c)}{a^2b^2}y + \tfrac{(b^2c^2+a^2s^2)x_c^2+2(b^2cs-a^2cs)x_cy_c+(b^2s^2+a^2c^2)y_c^2}{a^2b^2}-1\\
0&=\bigl(\tfrac{c^2}{a^2}+\tfrac{s^2}{b^2}\bigr)x^2+2cs\bigl(\tfrac{1}{a^2}-\tfrac{1}{b^2}\bigr)xy+\bigl(\tfrac{s^2}{a^2}+\tfrac{c^2}{b^2}\bigr)y^2-2\bigl[(\tfrac{c^2}{a^2}+\tfrac{s^2}{b^2}\bigr)x_c-\bigl(\tfrac{cs}{a^2}-\tfrac{cs}{b^2}\bigr)y_c\bigr]x\\
&-2\bigl[\bigl(\tfrac{cs}{a^2}-\tfrac{cs}{b^2}\bigr)x_c+\bigl(\tfrac{s^2}{a^2}+\tfrac{c^2}{b^2})y_c\bigr]y+\bigl(\tfrac{c^2}{a^2}+\tfrac{s^2}{b^2}\bigr)x_c^2+2\bigl(\tfrac{cs}{a^2}-\tfrac{cs}{b^2}\bigr)x_cy_c+\bigl(\tfrac{s^2}{a^2}+\tfrac{c^2}{b^2}\bigr)y_c^2-1 =0.
\end{align*}
Let
$
A=\tfrac{c^2}{a^2}+\tfrac{s^2}{b^2}$, $
B=cs\bigl(\tfrac{1}{a^2}-\tfrac{1}{b^2}\bigr)$, $C=\tfrac{s^2}{a^2}+\tfrac{c^2}{b^2}$. Then, we can show that
$D=-Ax_c-By_c$ and $ E=-Bx_c-Cy_c$, while 
$F= Ax_c^2+2Bx_cy_c+Cy_c^2-1.$

\subsection{The Geometry of Concentric Ellipses}\label{ApxGeoCon}
Here we demonstrates Eq. \eqref{Eq00}. Let the points $\mathbf{z}_{11}=(x_{11},y_{11})$ and $\mathbf{z}_{21}=(x_{21},y_{21})$ be two points that lie on the inner and the outer ellipses, respectively, then they satisfy
$  A_ix_{i1}^2 + 2B_ix_{i1} y_{i1} + C_iy_{i1}^2 + 2D_ix_{i1} + 2E_iy_{i1} + F_i =0$,  for $i=1, 2$. The relation between the geometric and algebraic parameter vectors for a single ellipse is explained in Section \ref{relation}. To know the relationship between these two parametric spaces for concentric ellipses, let us first assume $\bt_1=(A_1,B_1,C_1,D_1,E_1,F_1)^\top$ and $\bt_2=(A_2,B_2,C_2,D_2,E_2,F_2)^\top$. After some calculations
one can show that $(A_2,B_2,C_2,D_2,E_2)=\tfrac{1}{(1+\varepsilon)^2}(A_1,B_1,C_1,D_1,E_1)$. Recall that multiplying the ellipse equation by a nonzero constant yields the same ellipse, thus, if two ellipses share the same tilt angle and center, then they share the same parameters given in the appendix. Thus, we drop their indices, i.e., we will assume both ellipses have the same $(A,B,C,D,E)$. However, both ellipses differ by $F_i$, i.e., $F_2 \neq \frac{1}{(1+\varepsilon)^2}F_1$, which follows from the following observation:
$ F_2 =  A_2x_c^2+C_2y_c^2+2B_2x_cy_c - 1  \neq \tfrac{F_1}{(1+\varepsilon)^2}$. Thus, we must account for $F_1$ and $F_2$ in our parametric space.
\subsection{Derivation of Eq. \eqref{TETO}} \label{Apx.TETO}

\begin{proof}[\textbf{Derivation of Eq. \eqref{TETO}}] Using $\tbxi_{ij}^\top\btth = 0$, $(\btth^\top\tilde{\bN}_{ S}\btth)=(\btth^\top\tilde{\bN}_{ T}\btth)$. Thus, one can rewrite $(\btth^\top\E[\mathbf{\rm  T}]\btth)$ as
\begin{align*}
 \btth^\top\E[\boldsymbol{ T}]\btth& = \sigma^2\Bigl( \btth^\top\tilde{\bN}_{\rm S}\btth - \summ(\txi^\top\minv\txi)(\btth^\top\bV_0[\txi]\btth)\Bigr), \\
&= \sigma^2\Bigl( \btth^\top\tilde{\bN}_{\rm T}\btth - {\rm tr}\bigl[\minv\summ(\btth^\top \bV_0[\txi]\btth)\txi\txi^\top\bigr]\Bigr).
\end{align*}\vspace{-5mm}
\end{proof}
\subsection{Derivation of the second-order bias for (i) the LS, (ii) the O'Leary, (iii) the Taubin, and (iv) the Semi-hyper LS methods.}\label{BiasDerviation}

\begin{proof}[\textbf{(i) Derivation of the bias for the LS}] One can see that using
$\E[\boldsymbol{\Delta_2{\hat{\theta}}}_{\rm  LS}] = \minv \bigl(\frac{(\btth^\top\E[\boldsymbol{ T}]\btth)}{\boldsymbol{\tilde{\theta}}^\top\tilde{\bN}\btth}\tilde{\bN}_{\rm LS}\btth - \E[\boldsymbol{ T}]\boldsymbol{\tilde{\theta}}\bigr).$
and substituting $\tilde{\bN} = \boldsymbol{ I}_7$ yield
$\E[\boldsymbol{\Delta_2{\hat{\theta}}}_{\rm  LS}] = \minv \bigl(\frac{\btth^\top\E[\boldsymbol{ T}]\btth}{\boldsymbol{\tilde{\theta}}^\top\boldsymbol{ I_7}\btth}\boldsymbol{ I_7}\btth - \E[\boldsymbol{ T}]\boldsymbol{\tilde{\theta}}\bigr)$. Now, substituting $\E[\boldsymbol{ T}]\btth$ in Eq. \eqref{ETtheta} yields the bias of the Least Squares method
$\E[\boldsymbol{\Delta_2{\hat{\theta}}}_{ \rm LS}] = -\sigma^2\minv\bigl(\tilde{\bN}_{ T}\btth + (\boldsymbol{e}^\top\btth)\boldsymbol{\tilde{\xi_{c}}} - \tilde{\boldsymbol{\pi}}\bigr)$.
\end{proof}

\begin{proof}[\textbf{(ii) Derivation of the bias for the O'Leary method}]For the O'Leary method, we have $\tilde{\bN}=\tilde{\bN}_{ OL}$ as defined in Eq. \eqref{OlearyConstraintMatrix}. Substituting $\tilde{\bN}_{ OL}$ in Eq. \eqref{EDel2} gives us
$\E[\boldsymbol{\Delta_2\hat{\theta}}_{\rm OL}] = \minv \bigl(\frac{(\btth^\top\E[\boldsymbol{ T}]\btth)}{\boldsymbol{\tilde{\theta}}^\top\bN_{\rm OL}\btth}\tilde{\bN}_{ \rm OL}\btth -  \E[\boldsymbol{ T}]\boldsymbol{\tilde{\theta}}\bigr).
$
Substituting in Eq. \eqref{ETtheta} and Eq. \eqref{TETO} yields
\begin{align*}
\E[\boldsymbol{\Delta_2{\hat{\theta}}}_{ \rm OL}] &= \sigma^2\minv \Bigl(\frac{(\btth^\top\tilde{\bN}_{\rm T}\btth) - {\rm  tr}[\minv\tilde{\bM}']}{(\btth^\top\tilde{\bN}_{ \rm OL}\btth)}\tilde{\bN}_{ \rm OL}\btth - \left(\tilde{\bN}_{\rm T}\btth+(\boldsymbol{e}^\top\btth)\tilde{\boldsymbol{\xi}_c} - \tilde{\boldsymbol{\pi}}\right)\Bigr)  \\
&= \sigma^2\minv \Bigl(\tilde{\bN}_{ \rm T}\btth - \frac{{\rm tr}[\minv\tilde{\bM}']\tilde{\bN}_{\rm OL}\btth}{\btth^\top\tilde{\bN}_{\rm  OL}\btth} - \left(\tilde{\bN}_{ \rm T}\btth+(\boldsymbol{e}^\top\btth)\tilde{\boldsymbol{\xi}_c} -  \tilde{\boldsymbol{\pi}}\right)\Bigr).
\end{align*}
If we define $ p=\frac{{\rm tr}[\minv\tilde{\bM}']}{\btth^\top\tilde{\bN}_{\rm OL}\btth}$, then the bias of the O'Leary Method can be expressed as
$ \E[\boldsymbol{\Delta_2^{\perp}{\theta}}_{ \rm OL}]= -\sigma^2\minv \left(p\tilde{\bN}_{ \rm OL}\btth  +(\boldsymbol{e}^\top\btth)\tilde{\xi_c} - \tilde{\boldsymbol{\pi}}\right).
$
\end{proof}

\begin{proof}[\textbf{(iii) Derivation of the bias for the Taubin method}] Here we use the constraint matrix of Taubin, i.e., $\rb{\tilde{\bN}_{\rm T}=\summ \bV_0[\txi]}$ . Substituting Eq. \eqref{TETO} into the general bias equation given in Eq. \eqref{EDel2} gives us
$$\E[\boldsymbol{\Delta_2}\hbtheta_{\rm T}] = \minv \bigl(\frac{
(\tbtheta^\top\E[\bT]\tbtheta)}{\btth^\top \tilde{\bN}_{ \rm T}\btth
}\tilde{\bN}_{ \rm T}\btth - \E[\bT] \boldsymbol{\tilde{\theta}}\bigr)
= \sigma^2\minv \bigl(\frac{(\btth^\top\tilde{\bN}_{ \rm T}\btth) -{\rm  tr [\minv\tilde{\bM}']} }{\btth^\top\tilde{\bN}_{ \rm T}\btth}\tilde{\bN}_{ \rm T}\btth - \E[\bT]\btth\bigr).
$$
Using Eq. \eqref{ETtheta}  and the definition $q$ yield
\begin{align*}
\E[\boldsymbol{\Delta_2{\theta}}_{ \rm T}] &= \sigma^2\minv \Bigl(\frac{(\btth^\top\tilde{\bN}_{ \rm T}\btth) -  {\rm tr}[\minv\tilde{\bM}']}{\btth^\top\tilde{\bN}_{ \rm T}\btth}\tilde{\bN}_{ \rm T}\btth - \left(\tilde{\bN}_{ \rm T}\btth+(\boldsymbol{e}^\top\btth)\tilde{\boldsymbol{\xi}_c} - \tilde{\boldsymbol{\pi}}\right)\Bigr). \\
&= -\sigma^2\minv\Bigl(q\tilde{\bN}_{\rm T}\btth + (\boldsymbol{e}^\top\btth)\boldsymbol{\tilde{\xi_c}} -\tilde{\boldsymbol{\pi}}\Bigr).
\end{align*}\vspace{-5mm}
\end{proof}

\begin{proof}[\textbf{(iv) Derivation of the bias for the Semi-Hyper LS}]For Semi-Hyper method, one has
$
\E[\boldsymbol{\Delta_2{\hat{\theta}}_{\rm S}}] = -\sigma^2\minv \left(\frac{(\btth^\top\E[\bT]\btth) \bigl(\tilde{\bN}_{ \rm T} + 2S[\boldsymbol{\tilde{\xi}}_c\boldsymbol{e}^\top]\bigr)\btth }{\btth^\top((\tilde{\bN}_{\rm T} + 2S[\boldsymbol{\tilde{\xi}}_c\boldsymbol{e}^\top])\btth}- \E[\bT]\boldsymbol{\tilde{\theta}}\right)$. Using Eq. \eqref{ETtheta} and Eq. \eqref{TETO} leads to
$$
\E[\boldsymbol{\Delta_2{\hat{\theta}}_{\rm S}}]= -\sigma^2\minv \left( \frac{(\btth^\top\tilde{\bN}_{ \rm T}\btth)-{\rm tr}[\minv\tilde{\bM}']\bigr) (\tilde{\bN}_{ \rm T} + 2S[\boldsymbol{\tilde{\xi}}_c\boldsymbol{e}^\top])\btth}{\boldsymbol{\tilde{\theta}}^\top (\tilde{\bN}_{ \rm T} + 2S[\boldsymbol{\tilde{\xi}}_c\boldsymbol{e}^\top])\btth} -\left(\tilde{\bN}_{\rm T}\btth + (\boldsymbol{e}^\top\btth)\boldsymbol{\tilde{\xi}}_c - \tilde{\boldsymbol{\pi}}\right)\right).
$$
Since $\txi^\top\btth=0$, then by substituting $q=\frac{\rm  tr[\minv \boldsymbol{\tilde{\bM}'}]}{\btth^\top\tilde{\bN}_{\rm  T}\btth}$ in the preceding equation, the bias of the Semi-Hyper LS method becomes
\begin{align*}
\E[\boldsymbol{\Delta_2{\hat{\theta}}_{\rm S}}]&=-\sigma^2\minv \left(\frac{(\btth^\top\tilde{\bN}_{\rm T}\btth)-{ \rm tr}[\minv\tilde{\bM}']}{\btth^\top\tilde{\bN}_{ \rm T}\btth}\tilde{\bN}_{ \rm T}\btth + (\boldsymbol{e}^\top\btth)\boldsymbol{\tilde{\xi}}_c -\left(\tilde{\bN}_{\rm  T}\btth + (\boldsymbol{e}^\top\btth)\boldsymbol{\tilde{\xi}}_c - \tilde{\boldsymbol{\pi}}\right)\right)\\
&=-\sigma^2\minv \left((1-q)(\tilde{\bN}_{\rm T}\btth+(\boldsymbol{e}^\top\btth)\boldsymbol{\tilde{\xi}}_c) -\left(\tilde{\bN}_{ \rm T}\btth + (\boldsymbol{e}^\top\btth)\boldsymbol{\tilde{\xi}}_c -  \tilde{\boldsymbol{\pi}}\right)\right).
\end{align*}\vspace{-5mm}
\end{proof}

\end{document}